# In-context Learning and Induction Heads

AUTHORS

Catherine Olsson\*, Nelson Elhage\*, Neel Nanda\*, Nicholas Joseph†, Nova DasSarma†, Tom Henighan†, Ben Mann†, Amanda Askell, Yuntao Bai, Anna Chen, Tom Conerly, Dawn Drain, Deep Ganguli, Zac Hatfield-Dodds, Danny Hernandez, Scott Johnston, Andy Jones, Jackson Kernion, Liane Lovitt, Kamal Ndousse, Dario Amodei, Tom Brown, Jack Clark, Jared Kaplan, Sam McCandlish, Chris Olah‡

\* Core Research Contributor;   † Core Infrastructure Contributor;   ‡ Correspondence to colah@anthropic.com; Author contributions statement below.

AFFILIATION

Anthropic

PUBLISHED

Mar 8, 2022

**Abstract:** "Induction heads" are attention heads that implement a simple algorithm to complete token sequences like [A][B] … [A] → [B]. In this work, we present preliminary and indirect evidence for a hypothesis that induction heads might constitute the mechanism for the majority of all "in-context learning" in large transformer models (i.e. decreasing loss at increasing token indices). We find that induction heads develop at precisely the same point as a sudden sharp increase in in-context learning ability, visible as a bump in the training loss. We present six complementary lines of evidence, arguing that induction heads may be the mechanistic source of general in-context learning in transformer models of any size. For small attention-only models, we present strong, causal evidence; for larger models with MLPs, we present correlational evidence.

**We recommend reading this paper as an HTML article.**

As Transformer generative models continue to scale and gain increasing real world use, [1, 2, 3, 4, 5] addressing their associated safety problems becomes increasingly important. *Mechanistic interpretability* – attempting to reverse engineer the detailed computations performed by the model – offers one possible avenue for addressing these safety issues. If we can understand the internal structures that cause Transformer models to produce the outputs they do, then we may be able to address current safety problems more systematically, as well as anticipating safety problems in future more powerful models.[1]

In the past, mechanistic interpretability has largely focused on CNN vision models [6], but recently, we presented some very preliminary progress on mechanistic interpretability for Transformer language models [7]. Specifically, in our prior work we developed a mathematical framework for decomposing the operations of transformers, which allowed us to make sense of small (1 and 2 layer attention-only) models and give a near-complete account of how they function. Perhaps the most interesting finding was the *induction head*, a circuit whose function is to look back over the sequence for previous instances of the current token (call it `A`), find the token that came after it last time (call it `B`), and then predict that the same completion will occur again (e.g. forming the sequence `[A][B] … [A] → [B]`). In other words, induction heads "complete the pattern" by copying and completing sequences that have occurred before. Mechanically, induction heads in our models are implemented by a circuit of two attention heads: the first head is a "previous token head" which copies information from the previous token into the next token, while the second head (the actual "induction head") uses that information to find tokens preceded by the present token. For 2-layer attention-only models,[2] we were able to show precisely that induction heads implement this pattern copying behavior and appear to be the primary source of in-context learning.

Ultimately, however, our goal is to reverse-engineer frontier language models (which often contain hundreds of layers and billions or trillions of parameters), not merely 2-layer attention-only models. Unfortunately, both the presence of many layers, and the presence of MLPs, makes it much more difficult to mathematically pin down the precise circuitry of these models. However, a different approach is possible: by empirically observing, perturbing, and studying the learning process and the formation of various structures, we can try to assemble an *indirect* case for what might be happening mechanistically inside the network. This is somewhat similar to how a neuroscientist might gain understanding of how part of the brain functions by looking at neural development over time, studying patients with an injury to that part of the brain, perturbing brain function in animals, or looking at a select small number of relevant neurons.

In this paper, we take the first preliminary steps towards building such an indirect case. In particular, we present preliminary and indirect evidence for a tantalizing hypothesis: *that induction heads might constitute the mechanism for the actual majority of all in-context learning in large transformer models*. Specifically, the thesis is that there are circuits which have the same or similar mechanism to the 2-layer induction heads and which perform a "fuzzy" or "nearest neighbor" version of pattern completion, completing `[A*][B*] … [A] → [B]`, where `A* ≈ A` and `B* ≈ B` are similar in some space; and furthermore, that these circuits implement most in-context learning in large models.

The primary way in which we obtain this evidence is via discovery and study of a *phase change* that occurs early in training for language models of every size (provided they have more than one layer), and which is visible as a bump in the training loss. During this phase change, the majority of in-context learning ability (as measured by difference in loss between tokens early and late in the sequence) is acquired, and simultaneously induction heads form within the model that are capable of implementing fairly abstract and fuzzy versions of pattern completion. We study this connection in detail to try to establish that it is causal, including showing that if we perturb the transformer architecture in a way that causes the induction bump to occur in a different place in training, then the formation of induction heads *as well as* formation of in-context learning simultaneously move along with it.

Specifically, the paper presents six complementary lines of evidence arguing that induction heads may be the mechanistic source of general in-context learning in transformer models of any size:

- **Argument 1** *(Macroscopic co-occurence)*: Transformer language models undergo a "phase change" early in training, during which induction heads form and simultaneously in-context learning improves dramatically.

- **Argument 2** *(Macroscopic co-perturbation):* When we change the transformer architecture in a way that shifts whether induction heads can form (and when), the dramatic improvement in in-context learning shifts in a precisely matching way.

- **Argument 3** *(Direct ablation):*  When we directly "knock out" induction heads at test-time in small models, the amount of in-context learning greatly decreases.

- **Argument 4** (*Specific examples of induction head generality*): Although we define induction heads very narrowly in terms of copying literal sequences, we empirically observe that these same heads also appear to implement more sophisticated types of in-context learning, including highly abstract behaviors, making it plausible they explain a large fraction of in-context learning.

- **Argument 5** (*Mechanistic plausibility of induction head generality):* For small models, we can explain mechanistically how induction heads work, and can show they contribute to in-context learning. Furthermore, the actual mechanism of operation suggests natural ways in which it could be re-purposed to perform more general in-context learning.

- **Argument 6** (*Continuity from small to large models*): In the previous 5 arguments, the case for induction heads explaining in-context learning is stronger for small models than for large ones. However, many behaviors and data related to both induction heads and in-context learning are smoothly continuous from small to large models, suggesting the simplest explanation is that mechanisms are the same.

Together the claims establish a circumstantial case that induction heads *might* be responsible for the majority of in-context learning in state-of-the-art transformer models. We emphasize that our results here are only the beginnings of evidence for such a case, and that like any empirical or interventional study, a large number of subtle confounds or alternative hypotheses are possible – which we discuss in the relevant sections. But we considered these results worth reporting, both because future work could build on our results to establish the claim more firmly, and because this kind of indirect evidence is likely to be common in interpretability as it advances, so we'd like to establish a norm of reporting it even when it is not fully conclusive.

Finally, in addition to being instrumental for tying induction heads to in-context learning, the phase change may have relevance to safety in its own right. Neural network capabilities — such as multi-digit addition — are known to sometimes abruptly form or change as models train or increase in scale [8, 1], and are of particular concern for safety as they mean that undesired or dangerous behavior could emerge abruptly. For example reward hacking, a type of safety problem, can emerge in such a phase change [9]. Thus, studying a phase change "up close" and better understanding its internal mechanics could contain generalizable lessons for addressing safety problems in future systems. In particular, the phase change we observe forms an interesting potential bridge between the microscopic domain of interpretability and the macroscopic domain of scaling laws and learning dynamics.

The rest of the paper is organized as follows. We start by clarifying several key concepts and definitions, including in-context learning, induction heads, and a "per-token loss analysis" method we use throughout. We then present the 6 arguments one by one, drawing on evidence from analysis of 34 transformers over the course of training, including more than 50,000 attention head ablations (the data of which is shown in the Model Analysis Table). We then discuss some unexplained "curiosities" in our findings, as well as reviewing related work.

# Key Concepts

## In-context Learning

In modern language models, tokens later in the context are easier to predict than tokens earlier in the context. As the context gets longer, loss goes down. In some sense this is just what a sequence model is designed to do (use earlier elements in the sequence to predict later ones), but as the ability to predict later tokens from earlier ones gets better, it can increasingly be used in interesting ways (such as specifying tasks, giving instructions, or asking the model to match a pattern) that suggest it can usefully be thought of as a phenomenon of its own. When thought of in this way, it is usually referred to as **in-context learning**.[3]

Emergent in-context learning was noted in GPT-2 [10] and gained significant attention in GPT-3 [1]. Simply by adjusting a "prompt", transformers can be adapted to do many useful things without re-training, such as translation, question-answering, arithmetic, and many other tasks. Using "prompt engineering" to leverage in-context learning became a popular topic of study and discussion [11, 12].

At least two importantly different ways of conceptualizing and measuring in-context learning exist in the literature. The first conception, represented in Brown *et al.*, focuses on *few-shot learning* [1] of specific tasks. The model is prompted with several instances of some "task" framed in a next-token-prediction format (such as few-digit addition, or English-to-French translation). The second conception of in-context learning, represented in Kaplan *et al.* [13], focuses on observing the loss at different *token indices*, in order to measure how much better the model gets at prediction as it receives more context. The first conception can be thought of as a micro perspective (focusing on specific tasks), where as the second conception can be seen as a macro perspective (focusing on general loss, which on average correlates with these tasks).

The "few-shot learning" conception of in-context learning has tended to receive greater community attention. The ability to do many different tasks with one large model, even without further fine-tuning, is a notable change to the basic economics of model training. Moreover, it gives evidence of wide-ranging general capabilities and the ability to adapt on the fly, which nudges us to re-examine what it means for a model to "understand" or to "reason".

However, for the purposes of this work, we focus instead on the Kaplan *et al*. conception: *decreasing loss* at *increasing token indices*. We do so because it's a more general framing of the phenomenon than "few-shot learning". A drawback of this definition is it fails to isolate specific behaviors of interest. At the same time, it allows us to measure models' overall ability to learn on-the-fly from the context, without depending on our specific choices of "task". We'll also see that, starting from this definition, we are *also* able to study a couple classic few-shot learning examples (see Argument 4).

Throughout this work we compute a simple heuristic measure of in-context learning:

- **In-context learning score:** the loss of the 500th token in the context minus the average loss of the 50th token in the context, averaged over dataset examples.

We chose the 500th and 50th token indices somewhat arbitrarily. The 500th token is near the end of a length-512 context, and the 50th token is far enough into the context that some basic properties of the text have been established (such as language and document type) while still being near the start. We will also show that picking different numbers here does not change our conclusions.

Finally, it is worth noting that in-context learning is of potentially special relevance to safety. In-context learning makes it harder to anticipate how a model might behave after a long context. In the longer run, concepts such as mesa-optimization or inner-alignment [14] postulate that meaningful learning or optimization could occur at test time (without changing the weights). In-context learning would be an obvious future mechanism for such hidden optimization to occur, whether or not it does so today. Thus, studying in-context learning seems valuable for the future.

(See Related Work for more on in-context learning, and Discussion for more on the connection to safety.)

## Induction Heads

In our previous paper, we discovered a special kind of attention head – which we named *induction heads* – in two layer attention-only models. Induction heads are implemented by a circuit consisting of a pair of attention heads in different layers that work together to copy or complete patterns. The first attention head copies information from the previous token into each token. This makes it possible for the second attention head to attend to tokens based on what happened before them, rather than their own content. Specifically, the second head (which we call the "induction head") search for a previous place in the sequence where the present token `A` occurred and attends to the next token (call it `B`), copying it and causing the model to be more likely to output `B` as the next token. That is, the two heads working together cause the sequence `…[A][B]…[A]` to be more likely to be completed with `[B]`.

Induction heads are named by analogy to inductive reasoning. In inductive reasoning, we might infer that if `A` is followed by `B` earlier in the context, `A` is more likely to be followed by `B` again later in the same context. Induction heads crystallize that inference. They search the context for previous instances of the present token, attend to the token which would come next if the pattern repeated, and increase its probability. Induction heads attend to tokens that would be predicted by basic induction (over the context, rather than over the training data).

Notice that induction heads are implementing a simple algorithm, and are *not* memorizing a fixed table of n-gram statistics. The rule `[A][B] … [A] → [B]` applies regardless of what `A` and `B` are.[4] This means that induction heads can in some sense work *out of distribution,* as long as local statistics early in the context are representative of statistics later. This hints that they may be capable of more general and abstract behavior.

Our previous paper focused on a few explorations of induction heads, including showing that these heads occur in 2-layer attention-only models (but not 1-layer models); tracking down how they operate mechanistically as part of our mathematical decomposition of transformers; and making an eigenvalue-based test for detecting their presence. However, we were a bit vague on the exact *definition* of induction heads: it was more like we found a cluster of behaviors and mechanisms that tended to occur together, and called heads in that cluster "induction heads".

In this paper our goal is to provide evidence for something more expansive: that induction heads play a major role in *general* in-context learning (not just literal `[A][B]...[A]→[B]` copying), for *large* models and not just for small 2-layer attention only models.  To do this clearly and coherently, we need a more precise definition of induction heads. Mechanistic analysis of weights and eigenvalue analysis are much more complicated in large models with MLP's, so for this paper we choose to *define* induction heads by their narrow empirical sequence copying behavior (the `[A][B]...[A]→[B]`), and then attempt to *show* that they (1) also serve a more expansive function that can be tied to in-context learning, and (2) coincide with the mechanistic picture for small models.

Formally, **we define an induction head** as one which exhibits the following two properties[5] on a repeated random sequence[6] of tokens:

- **Prefix matching:** The head attends back to previous tokens that were followed by the current and/or recent tokens.[7] That is, it attends to the token which induction would suggest comes next.
- **Copying**: The head's output increases the logit corresponding to the attended-to token.

In other words, induction heads are any heads that empirically increase the likelihood of `[B]` given `[A][B]...[A]` when shown a repeated sequence of completely random tokens. An illustration of induction heads' behavior is shown here:

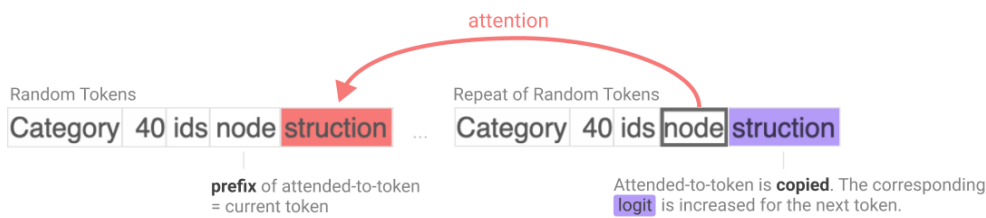

Note that, as a consequence, induction heads will tend to be good at repeating sequences wholesale. For example, given "The cat sat on the mat. The cat …", induction heads will promote the continuation "sat on the mat". This gives a first hint of how they might be connected to general in-context learning and even few-shot learning: they learn to repeat arbitrary sequences, which is a (simple) form of few-shot learning.

One of things we'll be trying to establish is that when induction heads occur in sufficiently large models and operate on sufficiently abstract representations, the very same heads that do this sequence copying *also* take on a more expanded role of *analogical sequence copying* or *in-context nearest neighbors*. By this we mean that they promote sequence completions like `[A*][B*] … [A] → [B]` where `A*` is not exactly the same token as `A` but similar in some embedding space, and also `B` is not exactly the same token as `B*`. For example, `A` and `A*` (as well as B and `B*`) might be the same word in different languages, and the induction head can then translate a sentence word by word by looking for "something like `A`", finding `A*` followed by `B*`, and then completing with "something like `B*`" (which is `B`). We are not yet able to prove mechanistically that induction heads do this in general, but in [Argument 4](#) we show empirical examples of induction heads behaving in this way (including on translation), and in [Argument 5](#) we point out that the known copying mechanism of induction heads in small models can be naturally adapted to function in this way.

## Per-Token Loss Analysis

To better understand how models evolve during training, we analyze what we call the "per-token loss vectors." The core idea traces back to a method used by Erhan *et al.* [15], and more generally to the idea of "function spaces" in mathematics.[8]

We start with a collection of models. (In our use, we'll train several different model architectures, saving dozens of "snapshots" of each over the course of training. We'll use this set of snapshots as our collection of models.) Next, we collect the log-likelihoods each model assigns to a consistent set of 10,000 random tokens, each taken from a different example sequence. We combine these log-likelihoods into a "per-token loss vector" and apply Principal Component Analysis (PCA):

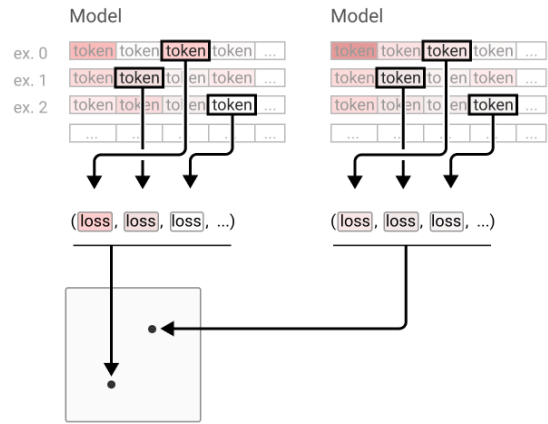

A more detailed discussion of technical details can be found in the Appendix.

By applying this method to snapshots over training for multiple models, we can visualize and compare how different models' training trajectories evolve in terms of their outputs. Since we're using PCA, each direction can be thought of as a vector of log-likelihoods that models are moving along. We particularly focus on the first two principal components, since we can easily visualize those. Of course, models also move in directions not captured by the first two principal components, but it's a useful visualization for capturing the highest-level story of training.

# Arguments that induction heads are the mechanism for the majority of in-context learning.

Now we'll proceed to the main part of the paper, which makes the case that induction heads may *provide the primary mechanism* for the majority of in-context learning for transformer models in general. As stated in the introduction, this is a very broad hypothesis and much of our evidence is indirect, but nevertheless we believe that all the lines of evidence together make a relatively strong, though not conclusive, case.

Before we go through the arguments, it's useful to delineate where the evidence is more conclusive vs. less conclusive. This is shown in the table below. For small, attention-only models, we believe we have strong evidence that attention heads are the mechanism for the majority of in-context learning, as we have evidence supported by ablations and mechanistic reverse engineering. Conversely, for all models, we can make a strong case that induction heads play *some role* in in-context learning, as we can demonstrate examples and show suggestive correlations. However, the larger the models get, the harder it is to establish that induction heads account for the actual *majority* of in-context learning. For large models with MLP's, we must therefore rely on mainly correlational evidence, which could be confounded. We explore alternate hypotheses throughout, including at the end of Argument 1 and again briefly in Argument 6.

SUMMARY OF EVIDENCE FOR SUB-CLAIMS (STRONGEST ARGUMENT FOR EACH)

|  | Small Attention-Only | Small with MLPs | Large Models |
|---|---|---|---|
| Contributes Some | Strong, Causal | Strong, Causal | Medium, Correlational & Mechanistic |
| Contributes Majority | Strong, Causal | Medium, Causal | Medium, Correlational |

Here is the list of arguments we'll be making, one per section, repeated from the introduction:

- **Argument 1** *(Macroscopic co-occurrence)***:** Transformer language models undergo a "phase change" early in training, during which induction heads form and simultaneously in-context learning improves dramatically.
- **Argument 2** *(Macroscopic co-perturbation):* When we change the transformer architecture in a way that shifts whether induction heads can form (and when), the dramatic improvement in in-context learning shifts in a precisely matching way.
- **Argument 3** *(Direct ablation):* When we directly "knock out" induction heads at test-time in small models, the amount of in-context learning greatly decreases.
- **Argument 4** (*Specific examples of induction head generality*): Although we define induction heads very narrowly in terms of copying literal sequences, we empirically observe that these same heads also appear to implement more sophisticated types of in-context learning, including highly abstract behaviors, making it plausible they explain a large fraction of in-context learning.
- **Argument 5** (*Mechanistic plausibility of induction head generality):* For small models, we can explain mechanistically how induction heads work, and can show they contribute to in-context learning. Furthermore, the actual mechanism of operation suggests natural ways in which it could be repurposed to perform more general in-context learning.
- **Argument 6** (*Continuity from small to large models*): In the previous 5 arguments, the case for induction heads explaining in-context learning is stronger for small models than for large ones. However, many behaviors and data related to both induction heads and in-context learning are smoothly continuous from small to large models, suggesting the simplest explanation is that mechanisms are the same.

For each argument, we'll have a similar table to the one in this section, showing the strength of the evidence provided by that claim as it applies to large/small models and some/most of context learning. The table above is the sum of the evidence from all six lines of reasoning.

# Argument 1: Transformer language models undergo a "phase change" during training, during which induction heads form and simultaneously in-context learning improves dramatically.

Our first line of evidence comes from correlating measures of in-context learning and measures of induction head presence over the course of training. Specifically, we observe a tight co-variation between them across dozens of models, of different sizes, trained on different datasets (See Model Analysis Table for more on the models in which we observe this co-occurrence.).

The table below summarizes the quality of this evidence for the models we have studied: it applies to both large and small models, and is the expected outcome if induction heads were responsible for the majority of in-context learning, but it is only correlational and so could be confounded (discussed more below).

STRENGTH OF ARGUMENT FOR SUB-CLAIMS

|  | Small Attention-Only | Small with MLPs | Large Models |
| --- | --- | --- | --- |
| Contributes Some | Medium, Correlational | Medium, Correlational | Medium, Correlational |
| Contributes Majority | Medium, Correlational | Medium, Correlational | Medium, Correlational |

Our first observation is that if we measure in-context learning for transformer models over the course of training (defined as the 50th token loss minus the 500th token loss as described in Key Concepts), it develops abruptly in a narrow window early in training (roughly 2.5 to 5 billion tokens) and then is constant for the rest of training (see figure below). Before this window there is less than 0.15 nats of in-context learning, after it there is roughly 0.4 nats, an amount that remains constant for the rest of training and is also constant across many different model sizes (except for the one layer model where not much in-context learning ever forms). This seems surprising – naively, one might expect in-context learning to improve gradually over training, and improve with larger model sizes,[9] as most things in machine learning do.

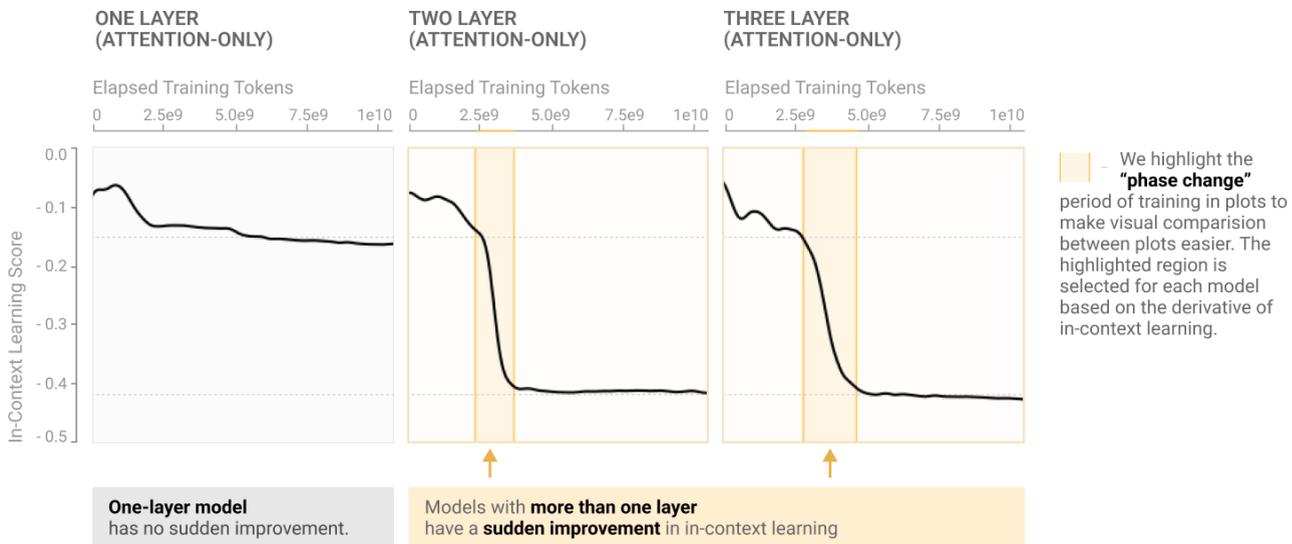

MODELS WITH MORE THAN ONE LAYER HAVE AN ABRUPT IMPROVEMENT IN IN-CONTEXT LEARNING

Although we only show three models above, the pattern holds true very generally: many examples are shown in the Model Analysis Table later in the paper, including models of varying model architecture and size.

One might wonder if the sudden increase is somehow an artifact of the choice to define in-context learning in terms of the difference between the 500th and 50th tokens. We'll discuss this in more depth later. But for now, an easy way to see that this is a robust phenomena is to look at the derivative of loss with respect to the logarithm token index in context. You can think of this as measuring something like "in-context learning per ε% increase in context length." We can visualize this on a 2D plot, where one axis is the amount of training that has elapsed, the other is the token index being predicted. Before the phase change, loss largely stops improving around token 50, but after the phase change, loss continues to improve past that point.

**DERIVATIVE OF LOSS WITH RESPECT TO LOG TOKEN INDEX**

The rate at which loss decreases with increasing token index can be thought of as something like "in-context learning per token". This appears to be most naturally measured with respect to the log number of tokens.

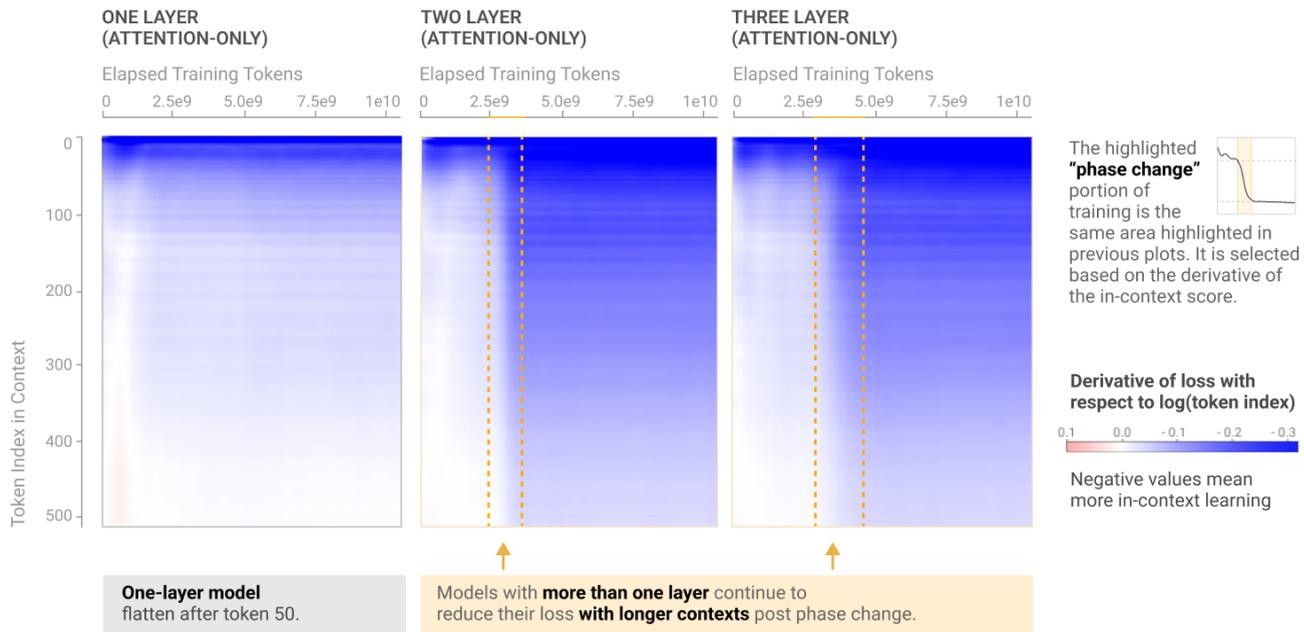

It turns out that a sudden improvement in in-context learning isn't the only thing that changes in this window. If we go through the attention heads of a model and and score them for whether they are induction heads (using a *prefix matching score* which measures their ability to perform the task we used to define induction heads in Key Concepts), we find that induction heads form abruptly during exactly the same window where in-context learning develops (figure below). Again we show only a few models here, but a full set is shown in the Model Analysis Table. The exception is the one-layer model, where induction heads never form – just as in-context learning never substantially develops for the one-layer model.

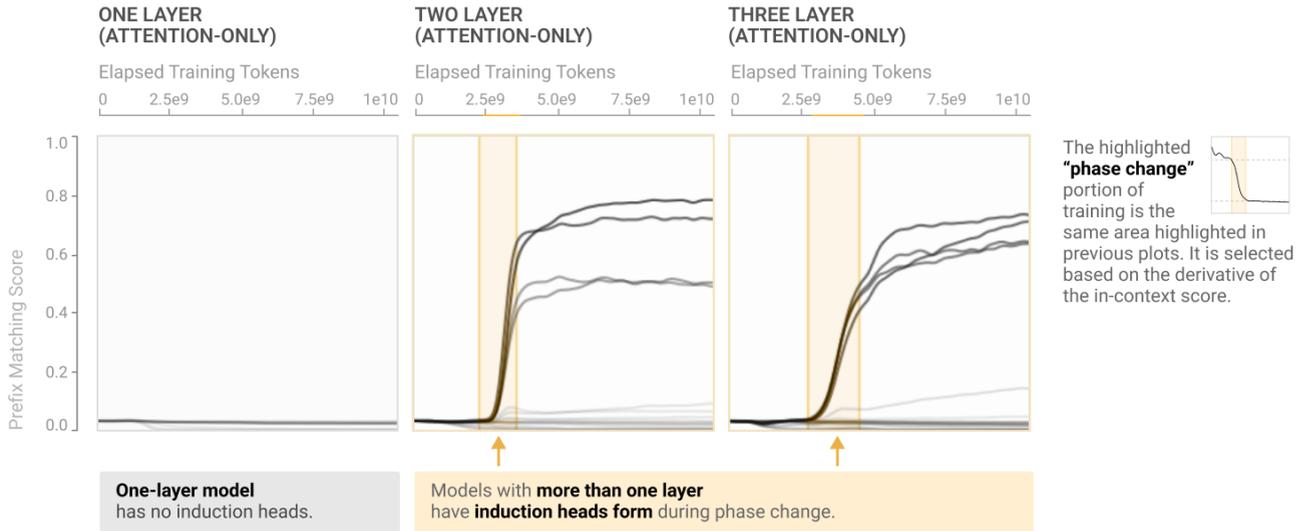

**INDUCTION HEADS FORM IN PHASE CHANGE**
Each line is an attention head, scored by the "prefix matching" evaluation introduced below.

**One-layer model** has no induction heads.

Models with **more than one layer** have **induction heads form** during phase change.

The highlighted **"phase change"** portion of training is the same area highlighted in previous plots. It is selected based on the derivative of the in-context score.

This already strongly suggests some connection between induction heads and in-context learning, but beyond just that, it appears this window is a pivotal point for the training process in general: whatever's occurring is visible as a bump on the training curve (figure below). It is in fact the only place in training where the loss is not convex (monotonically decreasing in slope).

That might not sound significant, but the loss curve is averaging over many thousands of tokens. Many behaviors people find interesting in language models, such as the emergence of arithmetic, would be microscopic on the loss curve. For something to be visible at that scale suggests it's a widespread, major change in model behavior. This shift also appears to be the first point where, at least for small models, the loss curve diverges from a one-layer model – which does not display the bump, just as it does not display the other abrupt changes.

## LOSS CURVES DIVERGE DURING PHASE CHANGE

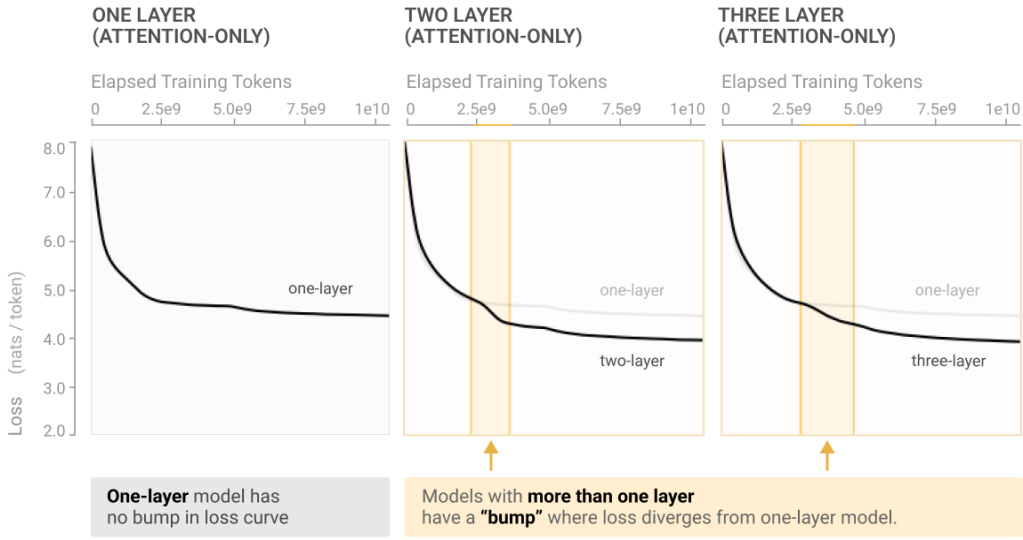

We can also apply principal components analysis (PCA) to the per-token losses, as described in per-token-loss analysis, which allows us to summarize the main dimensions of variation in how several models' predictions vary over the course of training.

Below we show the first two principal components of these models' predictions, with the golden outline highlighting the same interval shown above, when in-context learning abruptly improved. We see that the training trajectories pivot during exactly the same window where the other changes happen. In some sense, whatever is occurring when in-context learning improves is the primary deviation from the basic trajectory our transformers follow during the course of their training. Once again the only exception is the one-layer model – where induction heads cannot form and in-context learning does not improve.

## PER-TOKEN LOSS PRINCIPAL COMPONENT ANALYSIS

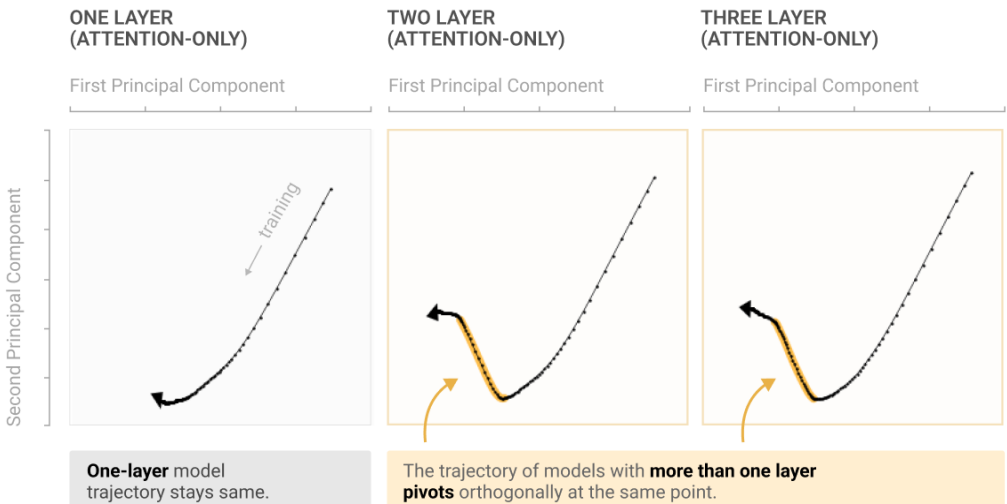
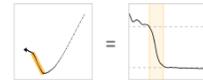

In summary, the following things all co-occur during the same abrupt window:

- Capacity for in-context learning sharply improves (as measured via the in-context learning score).
- Induction heads form.
- Loss undergoes a small "bump" (that is, the loss curve undergoes a period of visibly steeper improvement than the parts of the curve before and after).
- The model's trajectory abruptly changes (in the space of per-token losses, as visualized with PCA).

Collectively these results suggest that some important transition is happening during the 2.5e9 to 5e9 token window early in training (for large models this is maybe 1-2% of the way through training). We call this transition "the phase change", in that it's an abrupt change that alters the model's behavior and has both macroscopic (loss and in-context learning curves) and microscopic (induction heads) manifestations, perhaps analogous to e.g. ice melting.[10]

## Looking at the Phase Change More Closely

A natural explanation would be that for all these models, the induction heads *implement* in-context learning: their formation is what drives all the other changes observed. To strengthen this hypothesis a bit, we check a few things. First, the window where the phase change happens doesn't appear to correspond to a scheduled change in learning rate, warmup, or weight decay; there is not some known exogenous factor precipitating everything. Second, we tried out training some of the small models on a different dataset, and we observed the phase change develop in the same way (see Model Analysis Table for more details).[11]

Third, to strengthen the connection a little more, we look qualitatively and anecdotally at what's going on with the model's behavior during the phase change. One way to do this is to look at specific tokens the model gets better and worse at predicting. The model's loss is an average of billions of log-likelihood losses for individual tokens. By pulling them apart, we can get a sense for what's changed.

Concretely, let's pick a piece of text – for fun, we'll use the first paragraph of Harry Potter – and look at the differences in log-likelihoods comparing the start and the end of the phase change.[12] We'll notice that the majority of the changes occur when tokens are repeated multiple times in the text. If a sequence of tokens occurs multiple times, the model is better at predicting the sequence the second time it shows up. On the other hand, if a token is followed by a different token than it previously was, the post-phase-change model is worse at predicting it:

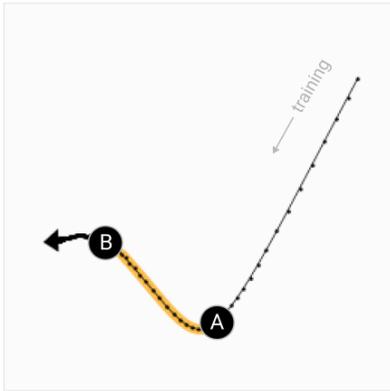

**PCA PLOT OF PER-TOKEN LOSSES** — **B** - **A** **PER-TOKEN LOSSES ON HARRY POTTER**

The training trajectory of models abruptly pivots during a "phase change." We look at the difference and loss before and after. As seen on the right, the change seems to be related to whether text is repeated. The qualitative change is similar across all models.

When text is repeated, the post phase-change model better predicts tokens:

… while cases where the same token is followed by a different token are given lower probability:

We can also do this same analysis over the course of model training. The loss curve is the average of millions of per-token loss curves. We can break this apart and look at the loss curves for individual tokens.

In particular, let's take a look at per-token loss trajectories for two tokens in the first paragraph of Harry Potter. In red, we show a token whose prediction gets dramatically *better* during the phase change: it's the last of four tokens in " The Durs**leys**", a sequence that appears several times in the text. In blue, we show a token that gets meaningfully *worse* during the phase change: it's the first-ever appearance of " Mrs Potter", after both previous instances of " Mrs" were followed by " Dursley" instead.

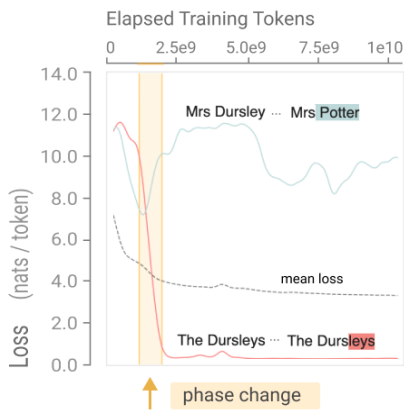

**PER-TOKEN LOSSES OVER TRAINING**

We look at two specific tokens in the Harry Potter passage a to highlight their movement during the phase change.

**Mrs Potter :** This token's loss increases during the phase change, because the model incorrectly predicts that the next token should be "D" from "Dursley".

**The Durs*leys* :** This token's loss decreases during the phase change, because the model correctly predicts "leys" based on a previous example.

All of this shows that during the phase change, we see exactly the behaviors we'd expect to see if induction heads were indeed contributing the majority of in-context learning.

## Assessing the Evidence

Despite all of the co-occurrence evidence above (and more in the [Model Analysis Table](#)), the fact remains that we haven't shown that induction heads are the primary mechanism for in-context learning. We have merely shown that induction heads form at the same time as in-context learning, and that in-context learning does not improve thereafter. There are a number of potential confounds in this story. Below we summarize reasons in favor and against thinking the connection is causal. The argument in favor might look like this:

- The formation of induction heads is correlated with a great increase in models' capacity for in-context learning, for a wide variety of models large and small.
- It's highly improbable these two sharp transitions would co-occur across so many models by pure chance, without any causal connection.
- There is almost surely some connection, and the simplest possibility is that induction heads are the primary mechanism driving the observed increase in in-context learning. (However, as discussed below, it could also be a confounding variable.)
- Since over 75% of final in-context learning forms in this window, one might naively believe this to be the amount of in-context learning that induction heads are responsible for.

However, the following issues and confounds suggest caution:

- In large models, we have low time resolution on our analysis over training. Co-occurrence when one only has 15 points in time is less surprising and weaker evidence.
- Perhaps other mechanisms form in our models at this point, that contribute to not only induction heads, but also other sources of in-context learning. (For example, perhaps the phase change is really the point at which the model learns how to compose layers through the residual stream, enabling both induction heads and potentially many other mechanisms that also require composition of multiple heads) Put another way, perhaps the co-occurrence is primarily caused by a shared latent variable, rather than direct causality from induction heads to the full observed change in in-context learning.
- The fact that in-context learning score is roughly constant (at 0.4 nats) after the phase change doesn't necessarily mean that the underlying mechanisms of in-context learning are constant after that point. In particular, the metric we use measures a *relative* loss between the token index 500 and 50, and we know that the model's performance at token 50 improves over training time. Reducing the loss a fixed amount from a lower baseline is likely harder, and so may be driven by additional mechanisms as training time goes on.[13][14]

One point worth noting here is that the argument that induction heads account for most in-context learning *at the transition point* of the phase change is more solid than the argument that they account for most in-context learning *at the end of training* – a lot could be changing during training even as the in-context learning score remains constant.

# Argument 2: When we change the transformer architecture in a way that shifts when induction heads form or whether they can form, the dramatic improvement in in-context learning shifts in a precisely matching way.

STRENGTH OF ARGUMENT FOR SUB-CLAIMS

|  | Small Attention-Only | Small with MLPs | Large Models |
|---|---|---|---|
| Contributes Some | Medium, Interventional | Medium, Interventional | Weak, Interventional |
| Contributes Majority | Medium, Interventional | Medium, Interventional | Weak, Interventional |

One thing that falls short about Argument 1 is that we're just observing induction heads and in-context learning co-vary; like any observational study, it is not as convincing as if we actively changed one thing and measured what happened to the other. In this section we do a more "interventional" experiment, in which we change the model architecture in a way that makes induction heads easier to form, and observe the effect on in-context learning. The change makes a bigger difference for smaller models so is more convincing there, but also carries some amount of weight for larger models (see table above).

To design our experiment, we start with the observation (noted in the previous section) that the phase change and the corresponding improvement in in-context learning only occurs in transformers with more than one layer. This is what we'd predict if induction heads were the mechanism for the majority of in-context learning: induction heads require a composition of attention heads, which is only possible with two or more layers.[15]

Of course, the observation about one-layer models is pretty weak evidence by itself. (One could imagine one-layer models being different from models with more layers in all sorts of ways!) But it suggests a more general line of attack. If induction heads are the mechanism behind the large improvement in in-context learning, that makes predictions about the minimum architectural requirements in order to achieve the observed improvement. For a standard transformer, the important thing is to have two attention layers. But that's only because the key vectors need to be a function of the attended token and the token before it.

We define a "smeared key" architecture with a very simple modification that should make it easy for transformers of any depth to express induction heads. In our modified models, for each head $h$, we introduce a trainable real parameter $\alpha^h$ which we use as $\sigma(\alpha^h) \in [0, 1]$ to interpolate between the key for the current token and previous token[16]:

$$k_j^h = \sigma(\alpha^h) k_j^h + (1 - \sigma(\alpha^h)) k_{j-1}^h$$

The hypothesis that induction heads are the primary mechanism of in-context learning predicts that the phase change will happen in one-layer models with this change, and perhaps might happen earlier in models with more layers. If they're one of several major contributing factors, we might expect some of the in-context learning improvement to happen early, and the rest at the same time as the original phase change.

The results (figure below) are in line with the predictions: when we use the smeared-key architecture, in-context learning does indeed form for one-layer models (when it didn't before), and it forms earlier for two-layer and larger models. More such results can be seen in the model analysis table.

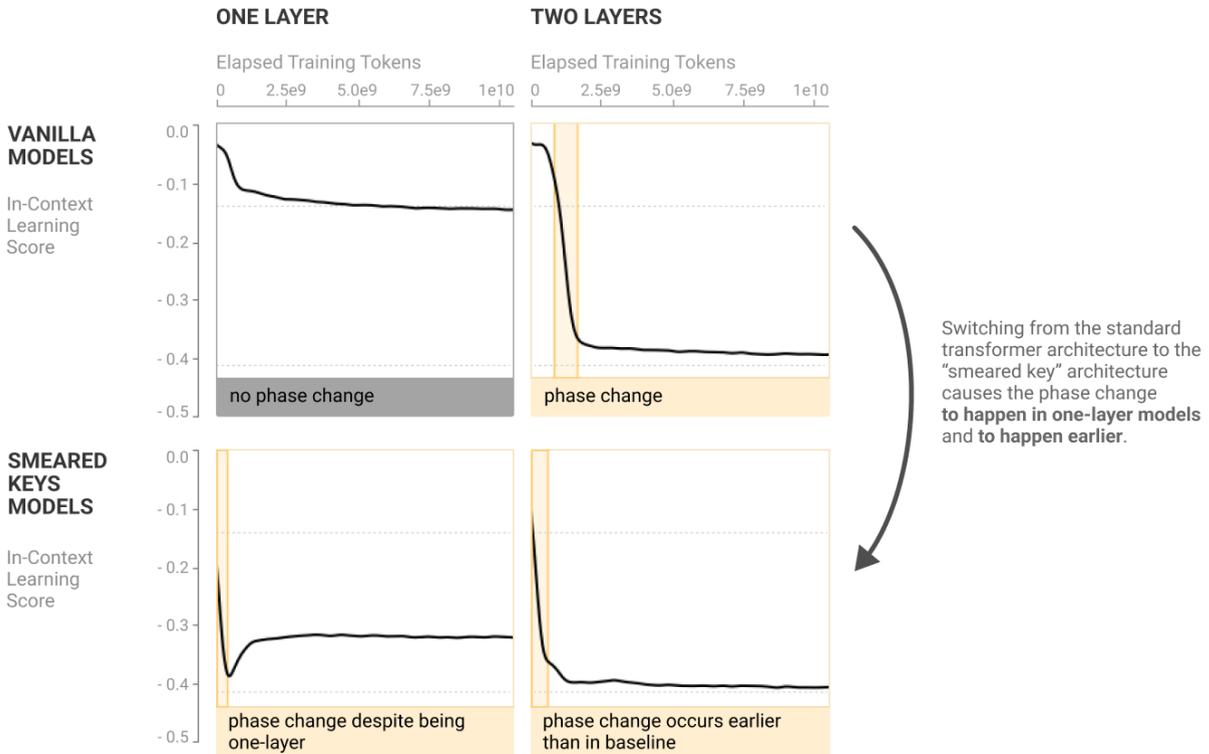

These plots are an excerpt of the Model Analysis Table; look there to see more analysis of how the smear models compare to their vanilla counterparts.

However, we probably shouldn't make too strong an inference about large models on this evidence. This experiment suggests that induction heads are the *minimal mechanism* for greatly increased in-context learning in transformers. But one could easily imagine that in larger models, this mechanism isn't the whole story, and also this experiment doesn't refute the idea of the mechanism of in-context learning changing over the course of training.

# Argument 3: When we directly "knock out" induction heads in small models at test-time, the amount of in-context learning greatly decreases.

STRENGTH OF ARGUMENT FOR SUB-CLAIMS

|  | Small Attention-Only | Small with MLPs | Large Models |
|---|---|---|---|
| Contributes Some | Strong, Causal | Strong, Causal |  |
| Contributes Majority | Strong, Causal | Medium, Causal |  |

For the cases they cover, ablations are by far our strongest evidence. The basic argument is that knocking out induction heads decreases the amount of in-context learning we observe in our models. By "knocking out" we mean that we remove a given attention head from the model at test time, doing a forward pass on the transformer without it. (See our methods section for exact details of how the ablation is done.)

The ablations presented below show how attention heads contribute to in-context learning, but we can also do ablations to study how attention heads contribute to the overall behavior change that occurs during the phase change (see the Model Analysis Table).

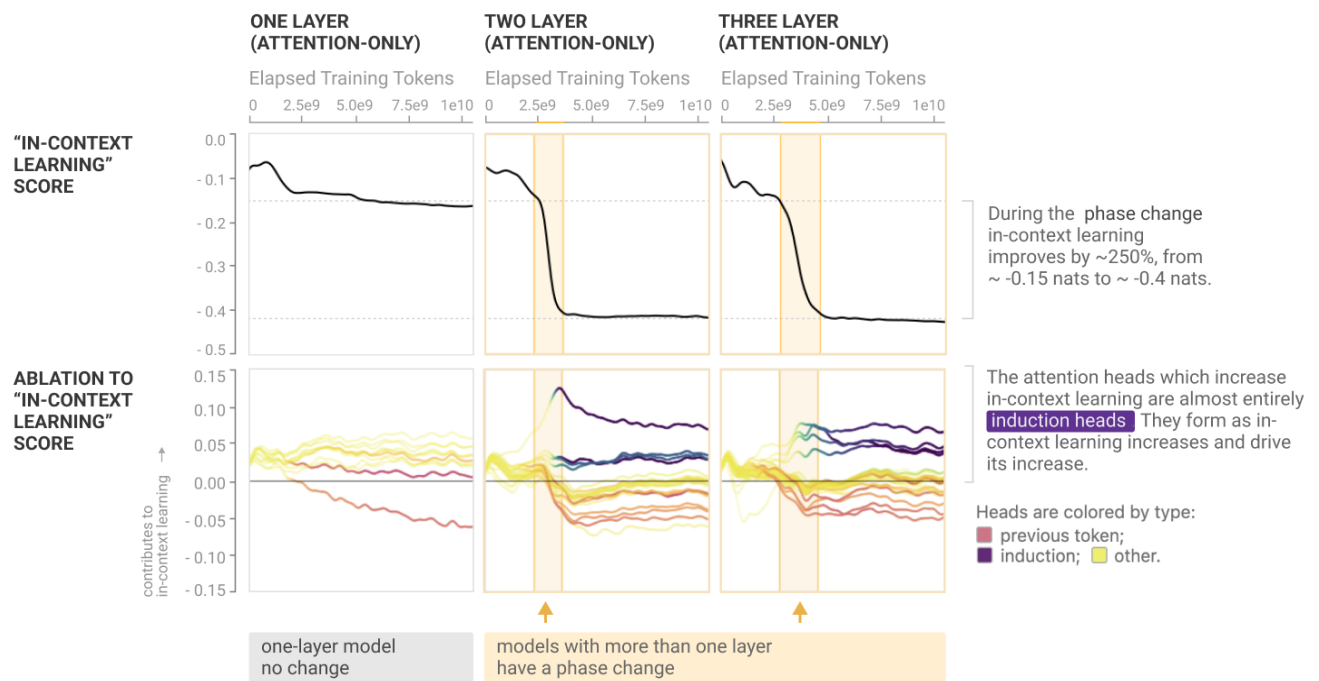

These plots are a small excerpt of the Model Analysis Table; look there to see the full evidence. See our methods section for exact details of how the ablation is done.

In fact, *almost all* the in-context learning in small attention-only models appears to come from these induction heads! This begins at the start of the phase change, and remains true through the end of training.[17]

Unfortunately, we do not have ablations for our full-scale models.[18] For the models where we do have ablations, it seems like this evidence is clearly dispositive that induction heads increase in-context learning (at least as we've chosen to evaluate it). But can we further infer that they're the primary mechanism? A couple considerations:

- In attention-only models, in-context learning must essentially be a sum of contributions from different attention heads.[19] But in models with MLPs, in-context learning could also come from interactions between MLP and attention layers. While ablating attention heads would affect such mechanisms, the relationship between the effect of the ablation on in-context learning and its true importance becomes more complicated.[20] As a result, we can't be fully confident that head ablations in MLP models give us the full picture.

- Our ablations measure the marginal effects of removing attention heads from the model. To the extent two heads do something similar and the layer norm before the logits rescales things, the importance of individual heads may be masked.

All things considered, we feel comfortable concluding from this that induction heads are the primary mechanism for in-context learning in small attention-only models, but see this evidence as only suggestive for the MLP case.

# Argument 4: Despite being defined narrowly as copying random sequences, induction heads can implement surprisingly abstract types of in-context learning.

STRENGTH OF ARGUMENT FOR SUB-CLAIMS

| | Small Attention-Only | Small with MLPs | Large Models |
|---|---|---|---|
| Contributes Some | | | Plausibility |
| Contributes Majority | | | Plausibility |

All of our previous evidence (in Arguments 1-3) focused on observing or perturbing the connection between induction head formation and macroscopic in-context learning. A totally different angle is to just find examples of induction heads implementing seemingly-difficult in-context learning behaviors; this would make it plausible that induction heads account for the majority of in-context learning. This evidence applies even to the very largest models (and we study up to 12B parameter models), but since it shows only a small number of tasks, it's only suggestive regarding in-context learning in general.

Recall that we <u>define</u> induction heads as heads that empirically copy arbitrary token sequences using a "prefix matching" attention pattern. Our goal is to find heads that meet this definition but *also* perform more interesting and sophisticated behaviors, essentially showing that induction heads in large models can be "generalizable".

In this argument, we show some anecdotal examples of induction heads from larger transformers (our 40-layer model with 13 billion parameters) that exhibit exactly such behaviors – namely literal copying, translation, and a specific type of abstract pattern matching. The behaviors are all of the form `[A*][B*]...[A][B]`, aka the "fuzzy nearest neighbor match" or "find something similar early in the sequence and complete the sequence in analogy". We verify that these heads score highly on our "copying" and "prefix matching" evaluations (that is, they increase the probability of the token they attend to, and attend to tokens where the prefix matches the present token on random text), and are thus "induction heads" by our strict empirical definition, at the same time as they *also* perform these more sophisticated tasks.

The results of some example heads are shown in the table below, and described in the subsections below.

| Head | Layer Depth | Copying score (?) | Prefix matching score (?) |
|---|---|---|---|
| Literal copying head | 21 / 40 | 0.89 | 0.75 |
| Translation head | 7 / 40 | 0.20 | 0.85 |
| Pattern-matching head | 8 / 40 | 0.69 | 0.94 |

## Behavior 1: Literal sequence copying

We'll start with the simplest case of a head that literally copies repeated text, to get familiar with the visualization interface we're using and the basic dynamics of these heads. We've selected an induction head which seems to perform very basic copying behavior and will look at how it behaves on the first paragraph of Harry Potter. We've repeated the first few sentences afterwards to show the head's behavior on longer segments of repeated text. **For the following interactive visualizations, we recommend visiting the HTML article.**

The visualization will show two different things:

- In red, "Attention" lets you see where the head is attending to predict the *next* token.
- In blue, "Logit attr" shows the earlier tokens that contributed to the prediction of the *current* token, using "direct-path" logit attribution.[21]

To start exploring the visualization, we suggest you visit the HTML article and try hovering your cursor over the second paragraph.

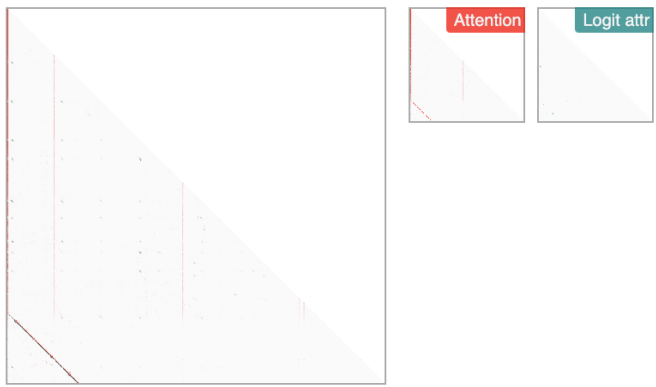

If you explore the visualization, you'll see that the head predicts repeating the names "Dursley" and "Potters"; the phrase "a small son"; and then the entire repeated sentences at the end. In all cases, these successful predictions are made by attending back to a previous instance where this phrase was present in the text.

## Behavior 2: Translation

It's a well-known result that language models can translate between languages. Intriguingly, we've encountered many examples of induction heads that can do translation. Here, we explore a head we found in layer 7 of our 40-layer model, showcasing translation between English, French, and German. (As with the others in this section, this head is *also* an "induction head" by the same definition we've been using all along, because when shown repeated *random* tokens, it uses a "prefix matching" attention pattern to copy the sequence verbatim.)

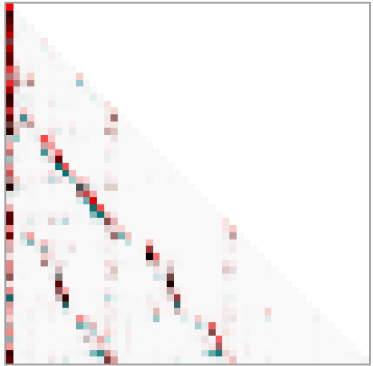
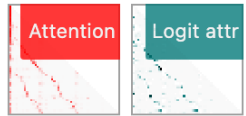

Note that the overall attention pattern (in red, top left) is more or less an "off-diagonal", but it meanders a little bit away from a sharp diagonal. The meandering is because different languages have different word order and token lengths. As this head attends sequentially to past tokens that will semantically come next, the attended token position in the earlier sentences jumps around.

The logit attribution patterns for this head are not perfectly sharp; that is, even in cases where the attention head is attending to the matching word in an earlier language, it does not always *directly* increase the logit of the corresponding prediction. We would guess that this is because this head's output needs to be further processed by later layers. However, taken overall, the direct logit attributions show clear evidence of contributing on net to the correct translation.

## Behavior 3: Pattern matching

In this final example, we show an attention head (found at layer 26 of our 40-layer model) which does more complex pattern matching. One might even think of it as learning a simple function in context! (Again, this head *also* scores highly on our measurements of "basic" induction behavior when shown repeated random sequences, so it is an induction head by that definition.)

To explore this behavior, we've generated some synthetic text which follows a simple pattern. Each line follows one of four templates, followed by a label for which template it is drawn from. The template is random selected, as are the words which fill in the template:

- (month) (animal): 0
- (month) (fruit): 1
- (color) (animal): 2
- (color) (fruit): 3

Below, we show how the attention head behaves on this synthetic example. To make the diagram easier to read, we've masked the attention pattern to only show the ":" tokens are the destination, and the logit attribution to only show where the output is the integer tokens.

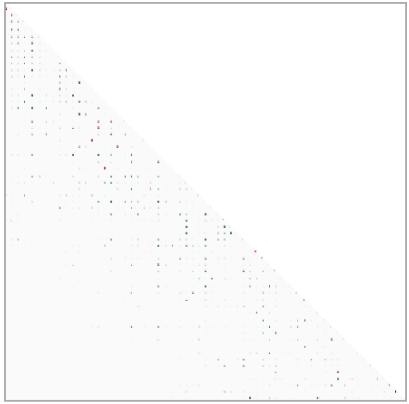
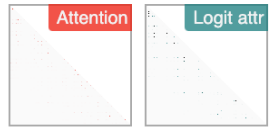

This head attends back to a previous instance of the correct category more often than not. It often knows to skip over lines where one of the words is identical but the pattern is wrong (such as "January bird" primarily attending to "April fish" and not "grey bird"). This head isn't perfect at this, but empirically it allocates about 65% of its attention from the colons to the correct positions, when tested on a range of similar problems.

## What's going on with more abstract heads that are also induction heads?

We emphasize again that the attention heads that we described above simultaneously implement *both* the abstract behaviors that we described, and these *very same* attention heads (as in, the exact same head in the same layer) *also* satisfy the formal definition of induction head (literal copying of random sequences using prefix matching). The comparison is not a metaphor or a blurring of the definition: induction heads which are *defined* by their ability to copy literal sequences *turn out to also* sometimes match more abstract patterns. This is what the table at the beginning of the section shows empirically.

But this still leaves the question: *why* do the same heads that inductively copy random text also exhibit these other behaviors? One hint is that these behaviors can be seen as "spiritually similar" to copying. Recall that where an induction head is defined as implementing a rule like `[A][B] … [A] → [B]`, our empirically observed heads also do something like `[A*][B*] … [A] → [B]` where `A*` and `B*` are similar to `A` and `B` in some higher-level representation. There are several ways these similar behaviors could be connected. For example, note that the first behavior is a special case of the second, so perhaps induction heads are implementing a more general algorithm that reverts to the special case of copying when given a repeated sequence [22]. Another possibility is that induction heads implement literal copying when they take a path through the residual stream that includes only them, but implement more abstract behaviors when they process the outputs of earlier layers that create more abstract representations (such as representations where the same word in English and French are embedded in the same place).

In Argument 5 we'll strengthen this argument by giving a mechanistic account of how induction heads (when doing simple copying with prefix-matching) attend back to the token that comes next in the pattern, and observe that the actual mechanism they use could naturally generalize to more abstract pattern matching. Our point in this section is just that it's actually quite natural for these more abstract induction heads to also exhibit the basic copying behaviors underlying our definition.

# Argument 5: For small models, we can explain mechanistically how induction heads work, and can show they contribute to in-context learning. Furthermore, the actual mechanism of operation suggests natural ways in which it could be re-purposed to perform more general in-context learning.

STRENGTH OF ARGUMENT FOR SUB-CLAIMS

|  | Small Attention-Only | Small with MLPs | Large Models |
|---|---|---|---|
| Contributes Some | Strong, Mechanistic | Strong, Mechanistic | Medium, Mechanistic |
| Contributes Majority | Weak, Mechanistic | | |

One of the main reasons we care about whether induction heads drive in-context learning is that we can understand them and so have a path to understanding in-context learning. But we can also turn this around: we can use our understanding of induction heads to make a purely logical argument that they *should* contribute to in-context learning.

We begin with a semi-empirical argument. Let's take for granted that induction heads behave the way we've described and empirically seen, searching the context for previous examples and copying what happened next. We should expect such a procedure to improve a model's ability to predict tokens later in its context. We're essentially using the previous context as data points for a nearest neighbor algorithm, and nearest neighbors improves as one gives it more data points. Therefore, if induction heads exist as described, they would contribute to in-context learning as we've defined it.

In some sense, this argument is quite strong if we're only arguing that there exist some cases where induction heads contribute to in-context learning. We've seen concrete examples above where induction heads improve token predictions by copying earlier examples. If nothing else, they must help in those cases! And more generally, our definition of induction heads (in terms of their behavior on repeated random sequences) suggests they behave this way quite generally. This argument doesn't say anything about *what fraction* of in-context learning is performed by induction heads, but it seems like a very strong argument that *some* is, both in large and small models.

But the really satisfying thing about this line of attack — namely, using our understanding of induction heads to anticipate their impact on in-context learning — is that we can actually drop the dependency on empirical observations of induction head behavior, at the cost of needing to make a more complex argument. In our previous paper, we were able to reverse engineer induction heads, showing from the parameter level how they implement induction behavior (and that they should). If we trust this analysis, we can know how induction heads behave without actually running them, and the argument we made in the previous paragraph goes through. Of course, there are some limitations. In the previous paper, we only reverse engineered a single induction head in a small attention-only model, although we can reverse engineered others (and have done so). A bigger issue is that right now we're unable to reverse engineer induction heads in models with MLP layers. But at least in some cases we've observed, we can look at the parameters of a transformer and identify induction heads, just as a programmer might identify an algorithm by reading through source code.

We can actually push this argument a bit further in the case of the two-layer attention only transformer we reverse engineered in the previous paper. Not only do we understand the induction heads and know that they should contribute to in-context learning, but there doesn't seem to really be an alternative mechanism that could be driving it.[23] This suggests that induction heads are the primary driver of in-context learning, at least in very small models.

The following section will briefly summarize reverse engineering induction heads. Note that it relies heavily on linking to our previous paper. *We do not expect it to be possible to follow without reading the linked portions.* After that, we briefly discuss how the described mechanism could also implement more abstract types of induction head behavior.

## Summary of Reverse Engineering Induction Heads

*Note: This section provides a dense summary with pointers to our previous paper; please see the previous paper for more information.*

Recall from Key Concepts that induction heads are defined as heads that exhibit both *copying* and *prefix matching*.

**Copying is done by the OV ("Output-Value") circuit.** One of the defining properties of an induction head is that it copies. Induction heads are not alone in this! Transformers seem to have quite a number of copying heads, of which induction heads are a subset. This is done by having a "copying matrix" OV circuit, most easily characterized by its positive eigenvalues.

**Prefix matching is implemented with K-composition (and to a lesser extent Q-composition) in the QK ("Query-Key") Circuit.** In order to do prefix matching, the key vector at the attended token needs to contain information about the *preceding* tokens — in fact, information about the *attended token itself* is quite irrelevant to calculating the attention pattern for induction.[24] In the models we study, "key shifting" occurs primarily using what we call K-composition. That is to say that an induction head's $W_K$ reads from a subspace written to by an earlier attention head. The most basic form of an induction head uses pure K-composition with an earlier "previous token head" to create a QK-Circuit term of the form $\text{Id} \otimes h_{prev} \otimes W$ where $W$ has positive eigenvalues. This term causes the induction head to compare the *current* token with every earlier position's *preceding* token and look for places where they're similar. More complex QK circuit terms can be used to create induction heads which match on more than just the preceding token.

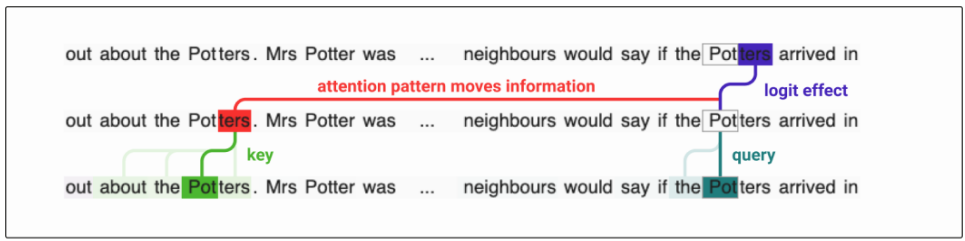
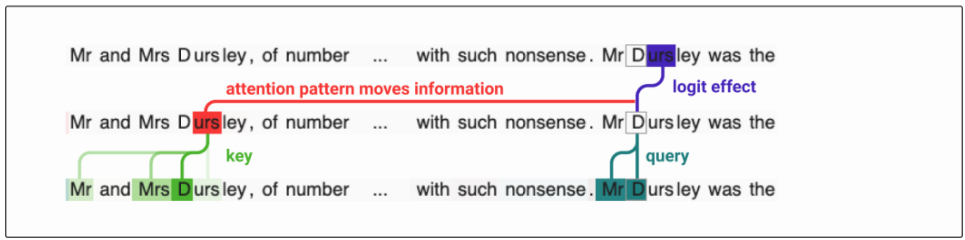

Induction heads use previous heads to shift key information and match it against the present token. As they get more sophisticated, they also shift query information.

**Combined, these are a detectable mechanism for induction heads.** In the small models we studied in our previous paper, all induction heads have the described QK term driving their attention pattern and a positive eigenvalue circuit performing copying.

**Some models use a different mechanism to implement induction heads.** In GPT-2 [10], we've seen evidence of a second "pointer-arithmetic" mechanism for induction heads. This mechanism makes use of the positional embedding and "Q-composition". In GPT-2, the earlier attention head attends to previous copies of the current token, and its $W_O V$ circuit copies their positional embedding into a subspace in the present token. The induction head then uses Q-composition to rotate that position embedding one token forward, and thereby attend to the following token. This mechanism isn't available to the models we study here, since they do not add positional information into the residual stream.[25]

## What About More Complex Induction Heads?

What about the induction heads we saw in Argument 2 with more complex behavior? Can we also reverse engineer them? Do they operate on the same mechanisms? Presently, fully reverse engineering them is beyond us, since they exist in large models with MLPs, which we don't have a strong framework for mechanistically understanding.  However, we hypothesize they're different in two ways: (1) using more complex QK terms rather than matching on just the previous token; and (2) matching and copying more abstract and sophisticated linguistic features, rather than precise tokens.

When we first introduced induction heads, we observed that they could be seen as a kind of "in-context nearest neighbor" algorithm. From this perspective, it seems natural that applying the same mechanism to more abstract features can produce more complex behavior.

# Argument 6:  Extrapolation from small models suggests induction heads are responsible for the majority of in-context learning in large models.

STRENGTH OF ARGUMENT FOR SUB-CLAIMS

|  | Small Attention-Only | Small with MLPs | Large Models |
|---|---|---|---|
| Contributes Some |  |  | Analogy |
| Contributes Majority |  |  | Analogy |

This argument is really an extended inference from all the above arguments. Arguments 1-5 present fairly strong evidence that in small transformers (especially small attention-only models), induction heads are responsible for the majority of in-context learning, while for large transformers, the evidence is not as strong. To what extent can one reasonably infer from small models that the same thing is happening in larger models? Obviously this is a matter of judgment.

The measurements in the model analysis table look fully analogous between the small attention-only, small models with MLPs, and full-scale model cases. Provided there's more than one layer, all of them go through a phase change. All of them have the same sharp increase in in-context learning, with the same rough amounts before and after the transition. All of them trace similar paths in PCA space. All of them form induction heads.

If things change from the small model case to the large model case, where do they change? And why is there no visible sign of the change in all our measurements?

On the flip side, there are many cases where large models behave very differently than small models (see discussion of phase changes with respect to model size in Related Work). Extrapolating from small models to models many orders of magnitude larger is something one should do with caution.

The most compelling alternative possibility we see is that other composition mechanisms may also form during the phase change. Larger models have more heads, which gives them more capacity for other interesting Q-composition and K-composition mechanisms that small models can't afford to express. If all "composition heads" form simultaneously during the phase change, then it's possible that above some size, non-induction composition heads could together account for more of the phase change and in-context learning improvement than induction heads do.

# Model Analysis Table

The arguments above are based on analysis of 34 decoder-only Transformer language models, with different snapshots saved over the course of training, for one run of training per model. The models are drawn from four different model series as follows:

- "*Small, attention-only models*", a series of models (from 1-layer to 6-layer) that do not have MLPs, and were trained specifically for the present investigations.
- "*Small models with MLPs*", a series of models (from 1-layer to 6-layer) that have both attention and MLP layers, and were trained specifically for the present investigations.
- *"Full-scale models",* a series of successively-larger models with MLPs (ranging from 4 layers and 13M parameters, up to 40 layers and 13B parameters), that are used as the basis for multiple projects at Anthropic.
- "*Smeared key models",* a targeted architectural experiment designed to allow transformers of any depth to express induction heads.

The dataset used for training the small models and smeared key models was an earlier version of the dataset described in Askell *et al*. [18], consisting of filtered common crawl data [19] and internet books, along with several other smaller distributions [20], including approximately 10% python code. The full-scale models were trained on an improved version of roughly the same data distribution. Additionally, another set of the small models was trained on a different dataset, consisting of just internet books, to explore the impact of varying the dataset. All models trained on a given dataset saw the same examples in the same order. Models never saw the same training data twice.

For more details on model architecture and training, continue past the table below, to the Model Details section.

In the Model Analysis Table below, each row includes a brief summary of the measurement shown. For a more in-depth explanation of the data collection and results analysis, see the Appendix.

# SMALL ATTENTION-ONLY TRANSFORMERS

This experiment series studies small *attention-only* transformers, because these are the models we have the strongest theoretical understanding of (see the framework paper). For these small models, we're able to capture snapshots with a high time resolution over early training. These experiments are the ones that most cleanly demonstrate the phase change and link it to induction heads. Note how (1) many models undergo a discontinuous phase change (seen most clearly in the PCA plot and in-context learning score); (2) it appears to be caused by induction heads (see ablations); and (3) the only model without a phase change is the one-layer model. More…

| ONE LAYER (ATTENTION-ONLY) | TWO LAYER (ATTENTION-ONLY) | THREE LAYER (ATTENTION-ONLY) | FOUR LAYER (ATTENTION-ONLY) | FIVE LAYER (ATTENTION-ONLY) | SIX LAYER (ATTENTION-ONLY) |

### PCA OF TOKEN LOSSES

The vector of per-token losses is a way to map different neural network behavior to the same vector space. We take 10,000 individual token predictions per model, and project them onto the first two principal components. This shows how the large-scale-behavior of multiple networks evolve over training. More…

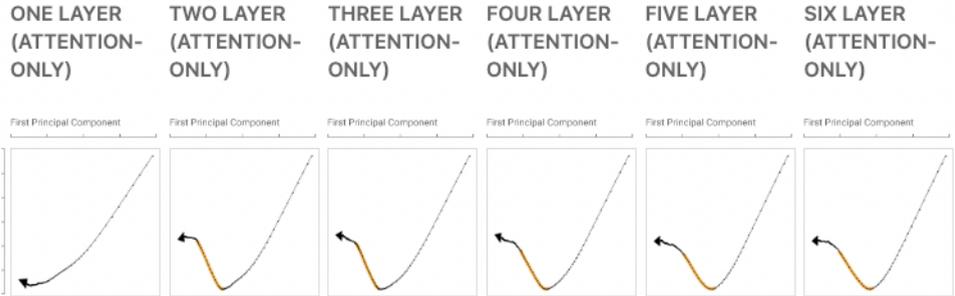

## MODELS OVER TRAINING

### LOSS OVER TRAINING

Loss curve showing $\mathrm{Loss}(n_{\mathrm{train}})$ after the model is trained on $n_{\mathrm{train}}$ tokens of training data. More…

The orange band ▯ denotes phase change interval.

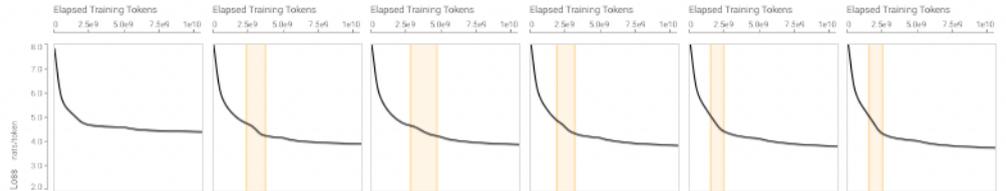

### LOSS OVER TRAINING, BROKEN DOWN BY CONTEXT INDEX

Heatmap of $\mathrm{Loss}(n_{\mathrm{train}}, i_{\mathrm{ctx}})$, the average loss of $i_{\mathrm{ctx}}$ token in context after $n_{\mathrm{train}}$ elapsed tokens of training. More…

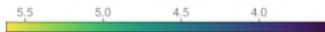

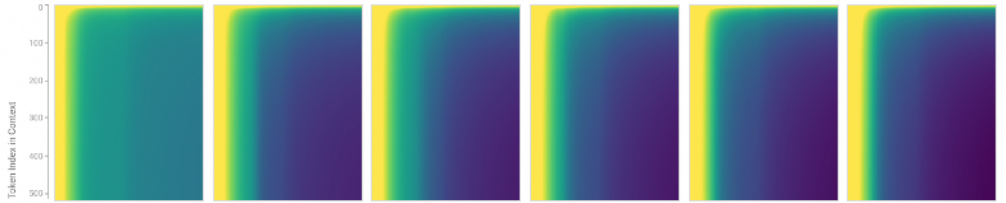

### DERIVATIVE OF LOSS WRT CONTEXT INDEX

Heatmap of $d\mathrm{Loss}(n_{\mathrm{train}}, i_{\mathrm{ctx}}) \,/\, d\ln(i_{\mathrm{ctx}})$. The log vertical partial derivative of the above graph. More…

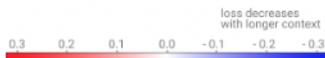

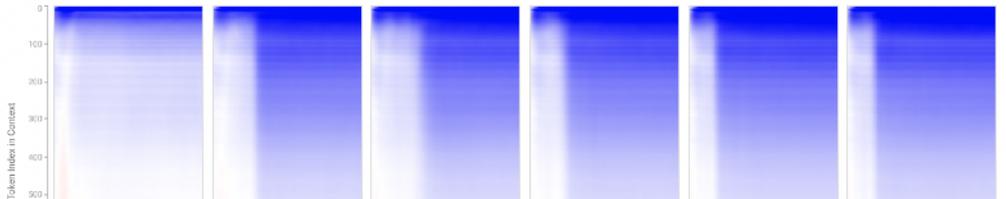

### "IN-CONTEXT LEARNING SCORE"

$\mathrm{Loss}(n_{\mathrm{train}}, i_{\mathrm{ctx}} = 500)$
$- \mathrm{Loss}(n_{\mathrm{train}}, i_{\mathrm{ctx}} = 50))$

How much better the model is at predicting the 500th token than the 50th? We use this as a proxy for how good the model is at in-context learning. More…

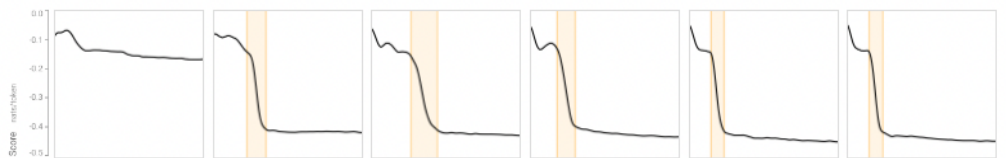

## ATTENTION HEADS OVER TRAINING

### "PREFIX MATCHING SCORE"

Each line is an attention head. Y axis is the average fraction of a head's attention weight given to the token we expect an induction head to attend to -- the token where the prefix matches the present context -- on random, synthetic data. Heads are colored by layer depth: 🟧 early; 🟦 late. More…

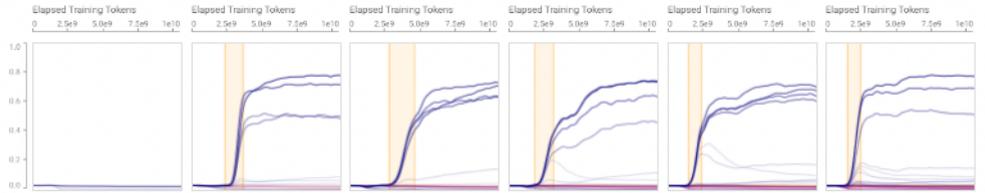

### TRACE OF QK EIGENVALUES

Each line is an attention head. Y axis is the trace of the $W$ in the attention head's $\mathrm{Id} \otimes h_{\mathrm{prev}} \otimes W$ QK circuit term, scaled by how well $h_{\mathrm{prev}}$ matches an ideal previous token head. Only defined for models more than one layer deep. Heads are colored by layer depth: 🟧 early; 🟦 late. More…

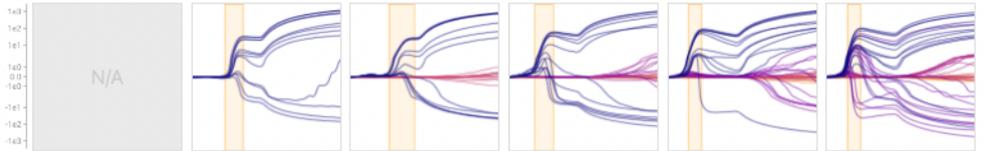

### ABLATION TO "BEFORE-AND-AFTER" VECTOR

Each line is an attention head. Y axis is the projection of the observed per-token loss differences when ablating an attention head onto the difference in per-token losses before and after the bump. More… Heads are colored by type: 🟥 previous token; 🟪 induction; 🟨 other.

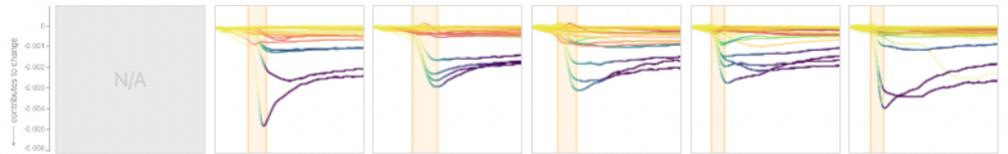

### ABLATION ATTRIBUTION TO "IN-CONTEXT LEARNING SCORE"

Each line is an attention head. Y axis is the change in the "in-context learning score" observed when ablating each attention head. More… Heads are colored by type: 🟥 previous token; 🟪 induction; 🟨 other.

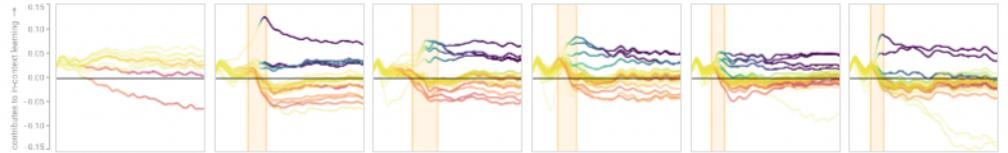

# SMALL TRANSFORMERS (WITH MLPS)

This experiment series studies small normal transformers (with MLP layers, unlike the above). For these small models, we're able to capture snapshots with a high time resolution over early training. The main observation to take away is that the results are essentially the same as the attention-only models (where we have a stronger theoretical understanding. More...

|  | ONE LAYER (WITH MLPS) | TWO LAYER (WITH MLPS) | THREE LAYER (WITH MLPS) | FOUR LAYER (WITH MLPS) | FIVE LAYER (WITH MLPS) | SIX LAYER (WITH MLPS) |

### PCA OF TOKEN LOSSES

The vector of per-token losses is a way to map different neural network behavior to the same vector space. We take 10,000 individual token predictions per model, and project them onto the first two principal components. This shows how the large-scale-behavior of multiple networks evolve over training. More...

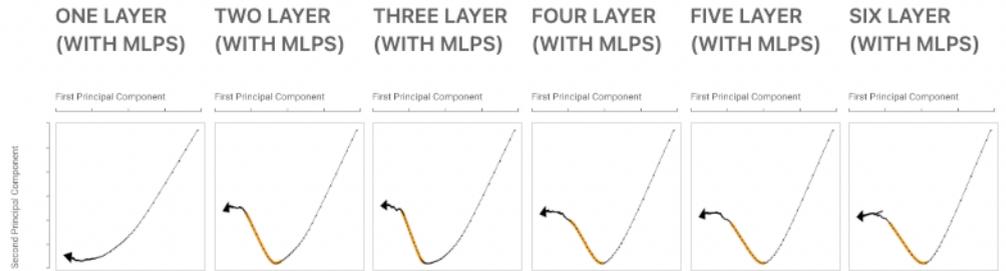

## MODELS OVER TRAINING

### LOSS OVER TRAINING

Loss curve showing $\text{Loss}(n_\text{train})$ after the model is trained on $n_\text{train}$ tokens of training data. More...

The orange band ▯ denotes phase change interval.

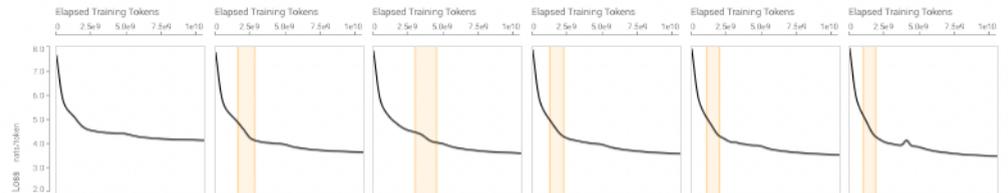

### LOSS OVER TRAINING, BROKEN DOWN BY CONTEXT INDEX

Heatmap of $\text{Loss}(n_\text{train}, i_\text{ctx})$, the average loss of $i_\text{ctx}$ token in context after $n_\text{train}$ elapsed tokens of training. More...

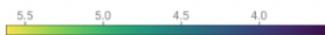

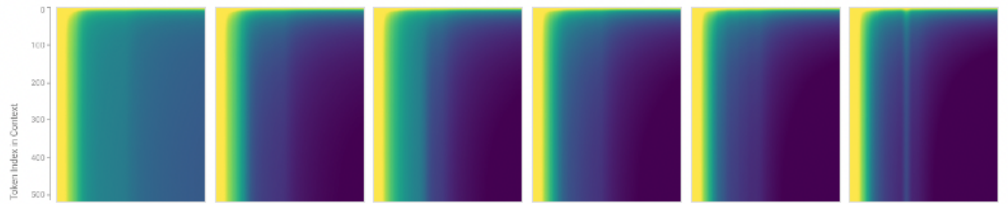

### DERIVATIVE OF LOSS WRT CONTEXT INDEX

Heatmap of $d\text{Loss}(n_\text{train}, i_\text{ctx}) \,/\, d\ln(i_\text{ctx})$. The log vertical partial derivative of the above graph. More...

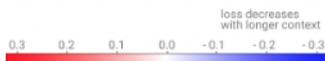

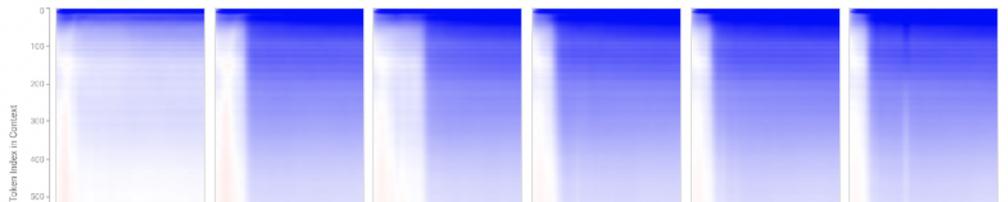

### "IN-CONTEXT LEARNING SCORE"

$\text{Loss}(n_\text{train}, i_\text{ctx} = 500)$
$- \text{Loss}(n_\text{train}, i_\text{ctx} = 50))$

How much better the model is at predicting the 500th token than the 50th? We use this as a proxy for how good the model is at in-context learning. More...

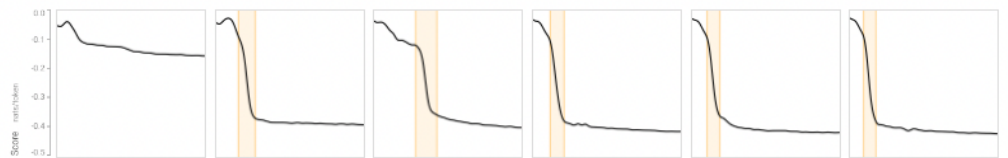

## ATTENTION HEADS OVER TRAINING

### "PREFIX MATCHING SCORE"

Each line is an attention head. Y axis is the average fraction of a head's attention weight given to the token we expect an induction head to attend to -- the token where the prefix matches the present context -- on random, synthetic data. Heads are colored by layer depth: 🟧 early; 🟦 late. More…

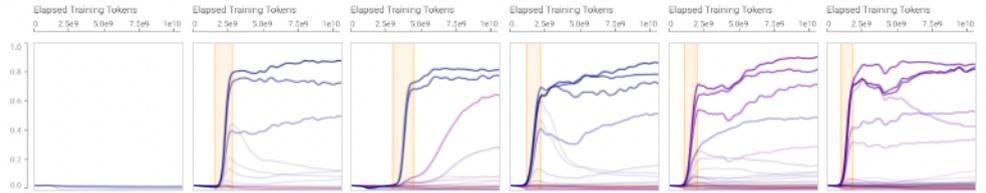

### TRACE OF QK EIGENVALUES

Each line is an attention head. Y axis is the trace of the $W$ in the attention head's $\mathrm{Id} \otimes h_{\mathrm{prev}} \otimes W$ QK circuit term, scaled by how well $h_{\mathrm{prev}}$ matches an ideal previous token head. Only defined for models more than one layer deep. Heads are colored by layer depth: 🟧 early; 🟦 late. More…

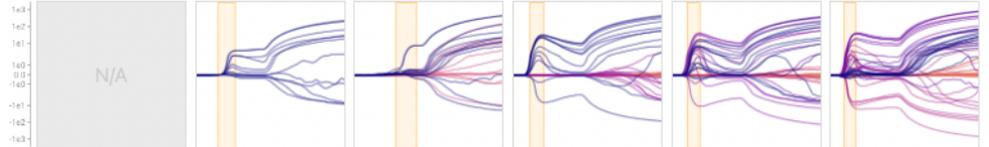

### ABLATION TO "BEFORE-AND-AFTER" VECTOR

Each line is an attention head. Y axis is the projection of the observed per-token loss differences when ablating an attention head onto the difference in per-token losses before and after the bump. More… Heads are colored by type: 🟥 previous token; 🟪 induction; 🟨 other.

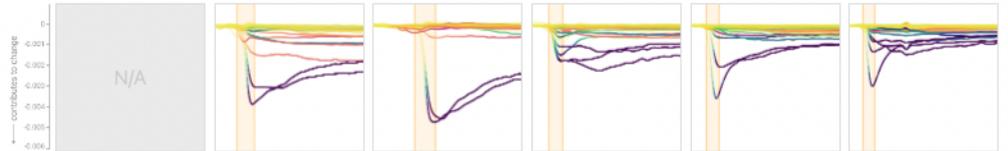

### ABLATION ATTRIBUTION TO "IN-CONTEXT LEARNING SCORE"

Each line is an attention head. Y axis is the change in the "in-context learning score" observed when ablating each attention head. More… Heads are colored by type: 🟥 previous token; 🟪 induction; 🟨 other.

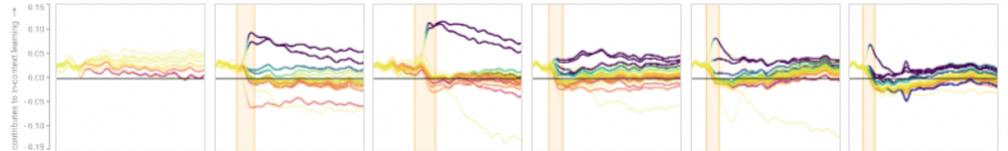

# FULL-SCALE TRANSFORMERS

This experiment series analyzes a sweep of typical transformers, spaced exponentially in size, up to a 13 billion parameter model. The goal is to show that the phenomena we observed in small models can be seen in larger models as well. For these models, we unfortunately have lower time resolution (note that the x-axis of most plots has changed to a log scale). We are also unable to provide complete ablations. The main observation to make is that results are roughly the same as seen in the small models. More...

4-LAYER TRANSFORMER (13M PARAMS) | 6-LAYER TRANSFORMER (42M PARAMS) | 10-LAYER TRANSFORMER (200M PARAMS) | 16-LAYER TRANSFORMER (810M PARAMS) | 24-LAYER TRANSFORMER (2.7B PARAMS) | 40-LAYER TRANSFORMER (13B PARAMS)

## PCA OF TOKEN LOSSES

The vector of per-token losses is a way to map different neural network behavior to the same vector space. We take 10,000 individual token predictions per model, and project them onto the first two principal components. This shows how the large-scale-behavior of multiple networks evolve over training. More...

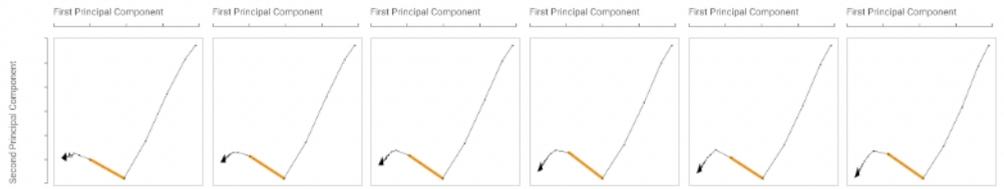

## MODELS OVER TRAINING

### LOSS OVER TRAINING

Loss curve showing $\text{Loss}(n_{\text{train}})$ after the model is trained on $n_{\text{train}}$ tokens of training data. More...

The orange band denotes phase change interval.

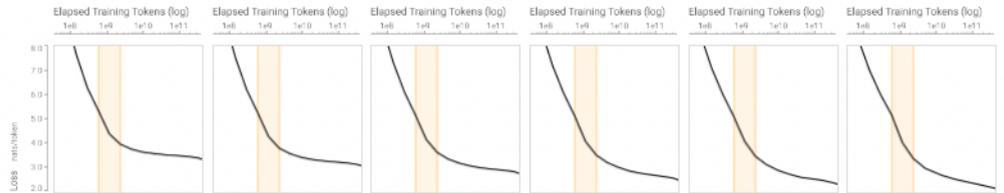

### LOSS OVER TRAINING, BROKEN DOWN BY CONTEXT INDEX

Heatmap of $\text{Loss}(n_{\text{train}}, i_{\text{ctx}})$, the average loss of $i_{\text{ctx}}$ token in context after $n_{\text{train}}$ elapsed tokens of training. More...

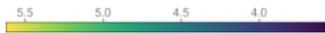

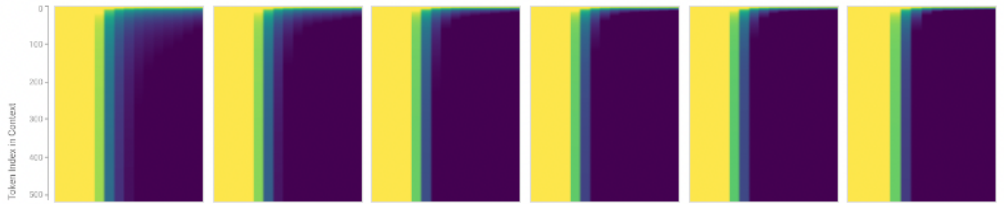

### DERIVATIVE OF LOSS WRT CONTEXT INDEX

Heatmap of $d\text{Loss}(n_{\text{train}}, i_{\text{ctx}}) \ / \ d\ln(i_{\text{ctx}})$. The log vertical partial derivative of the above graph. More...

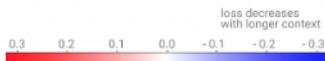

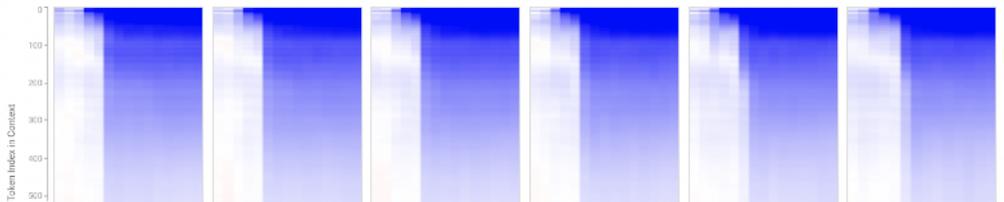

### "IN-CONTEXT LEARNING SCORE"

$\text{Loss}(n_{\text{train}}, i_{\text{ctx}} = 500)$
$- \text{Loss}(n_{\text{train}}, i_{\text{ctx}} = 50))$

How much better the model is at predicting the 500th token than the 50th? We use this as a proxy for how good the model is at in-context learning. More...

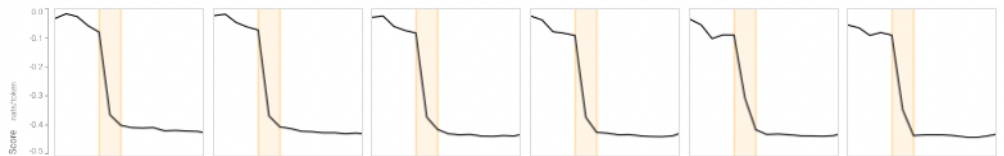

## ATTENTION HEADS OVER TRAINING

### "PREFIX MATCHING SCORE"

Each line is an attention head. Y axis is the average fraction of a head's attention weight given to the token we expect an induction head to attend to -- the token where the prefix matches the present context -- on random, synthetic data. Heads are colored by layer depth: ■ early; ■ late. More...

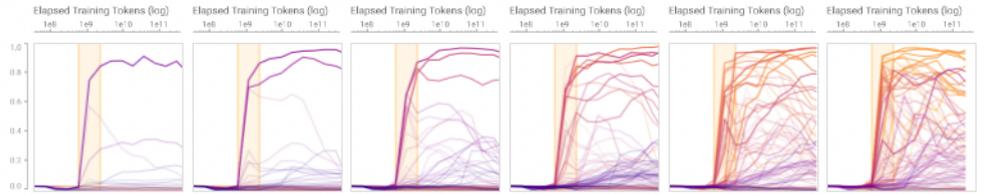

# "SMEARED KEY" ARCHITECTURE MODIFICATION + CONTROLS

This experiment series analyzes "smeared key" models, which are designed so that a single attention layer can implement induction heads (see Argument 2). For each smeared key model, a regular control model is provided (trained exactly the same) for comparison. The key observations are that (1) a one-layer smeared-key model experiences the phase change, while a regular one-layer model does not; and (2) the two-layer smeared-key model experiences a phase change earlier than a baseline model. More…

|  | ONE LAYER (VANILLA) | ONE LAYER (SMEARED KEYS) | TWO LAYER (VANILLA) | TWO LAYER (SMEARED KEYS) |

### LOSS OVER TRAINING

Loss curve showing $\text{Loss}(n_{\text{train}})$ after the model is trained on $n_{\text{train}}$ tokens of training data. More…

The orange band  denotes phase change interval.

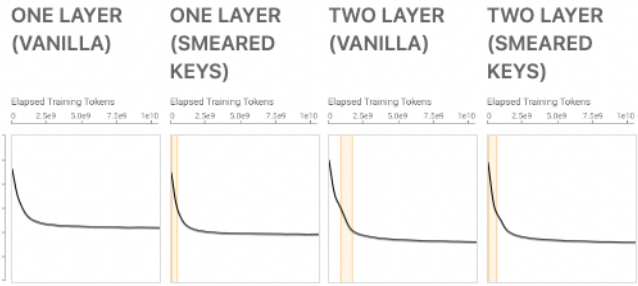

### LOSS OVER TRAINING, BROKEN DOWN BY CONTEXT INDEX

Heatmap of $\text{Loss}(n_{\text{train}}, i_{\text{ctx}})$, the average loss of $i_{\text{ctx}}$ token in context after $n_{\text{train}}$ elapsed tokens of training. More…

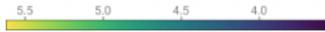

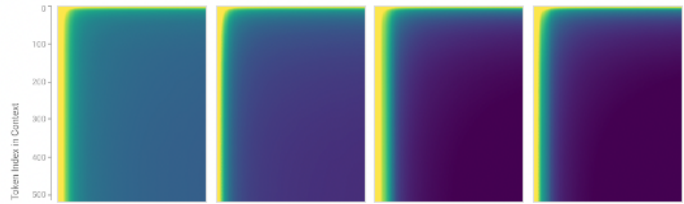

### DERIVATIVE OF LOSS WRT CONTEXT INDEX

Heatmap of $d\text{Loss}(n_{\text{train}}, i_{\text{ctx}}) \;/\; d\ln(i_{\text{ctx}})$. The log vertical partial derivative of the above graph. More…

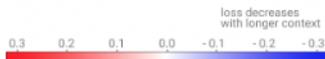

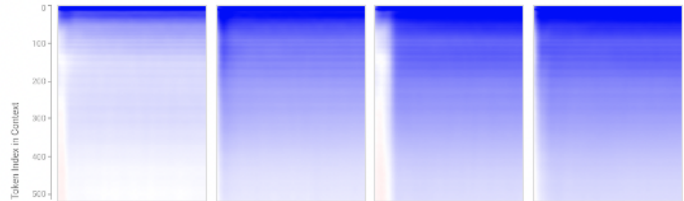

### "IN-CONTEXT LEARNING SCORE"

$\text{Loss}(n_{\text{train}}, i_{\text{ctx}} = 500)$
$- \text{Loss}(n_{\text{train}}, i_{\text{ctx}} = 50))$

How much better the model is at predicting the 500th token than the 50th? We use this as a proxy for how good the model is at in-context learning. More…

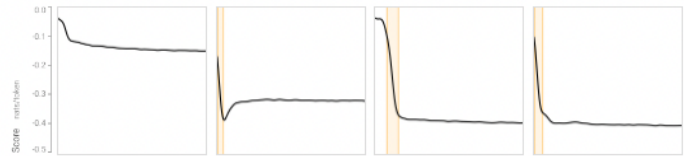

# SMALL ATTENTION-ONLY TRANSFORMERS (DIFFERENT DATASET)

This experiment series studies transformers of the same architecture as the small *attention-only* transformers, on a different dataset consisting only of internet books. More...

|  | ONE LAYER (ATTENTION-ONLY) | TWO LAYER (ATTENTION-ONLY) | THREE LAYER (ATTENTION-ONLY) | FOUR LAYER (ATTENTION-ONLY) | FIVE LAYER (ATTENTION-ONLY) | SIX LAYER (ATTENTION-ONLY) |

### LOSS OVER TRAINING

Loss curve showing $\text{Loss}(n_\text{train})$ after the model is trained on $n_\text{train}$ tokens of training data. More...

The orange band  denotes phase change interval.

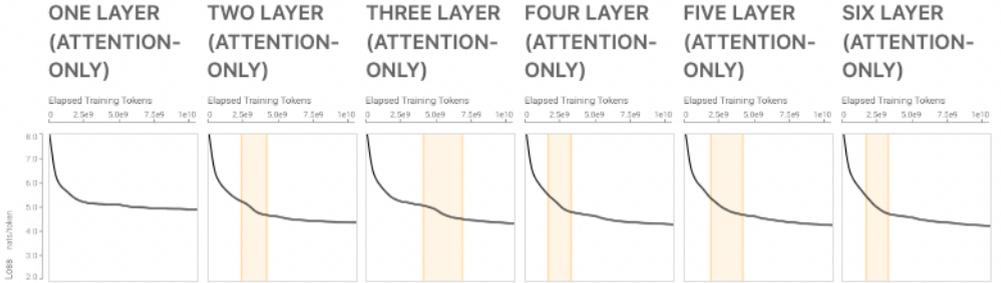

### LOSS OVER TRAINING, BROKEN DOWN BY CONTEXT INDEX

Heatmap of $\text{Loss}(n_\text{train}, i_\text{ctx})$, the average loss of $i_\text{ctx}$ token in context after $n_\text{train}$ elapsed tokens of training. More...

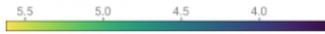

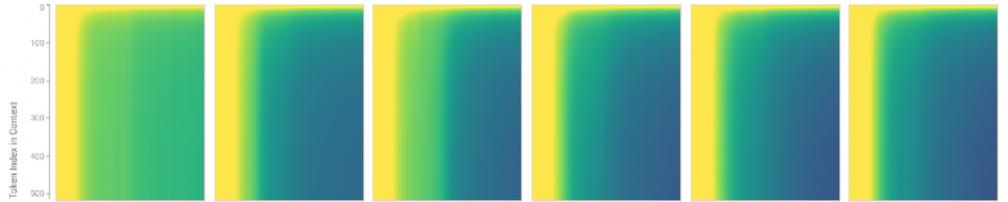

### DERIVATIVE OF LOSS WRT CONTEXT INDEX

Heatmap of $d\text{Loss}(n_\text{train}, i_\text{ctx}) \ / \ d\ln(i_\text{ctx})$. The log vertical partial derivative of the above graph. More...

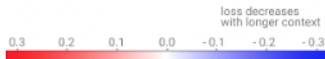

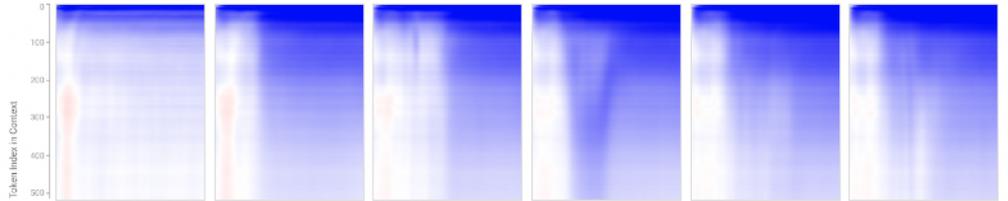

### "IN-CONTEXT LEARNING SCORE"

$\text{Loss}(n_\text{train}, i_\text{ctx} = 500)$
$- \text{Loss}(n_\text{train}, i_\text{ctx} = 50))$

How much better the model is at predicting the 500th token than the 50th? We use this as a proxy for how good the model is at in-context learning. More...

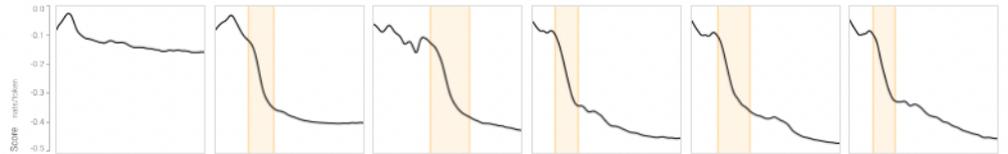

# SMALL TRANSFORMERS (WITH MLPS) (DIFFERENT DATASET)

This experiment series studies small normal transformers (with MLP layers), on a different dataset consisting only of internet books.
More...

| | ONE LAYER (WITH MLPS) | TWO LAYER (WITH MLPS) | THREE LAYER (WITH MLPS) | FOUR LAYER (WITH MLPS) | FIVE LAYER (WITH MLPS) | SIX LAYER (WITH MLPS) |
|---|---|---|---|---|---|---|

**LOSS OVER TRAINING**

Loss curve showing $\text{Loss}(n_{\text{train}})$ after the model is trained on $n_{\text{train}}$ tokens of training data.
More...

The orange band  denotes phase change interval.

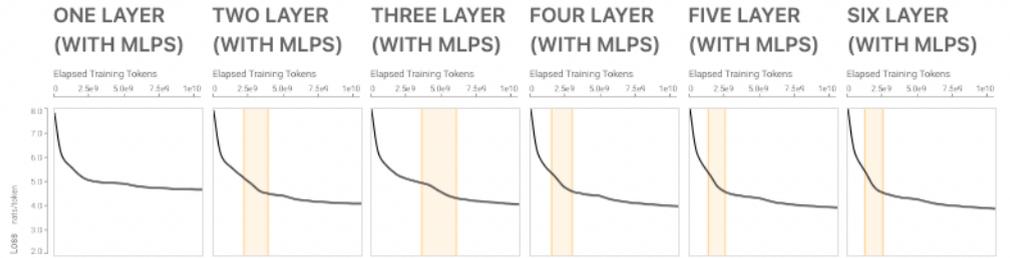

**LOSS OVER TRAINING, BROKEN DOWN BY CONTEXT INDEX**

Heatmap of $\text{Loss}(n_{\text{train}}, i_{\text{ctx}})$, the average loss of $i_{\text{ctx}}$ token in context after $n_{\text{train}}$ elapsed tokens of training. More...

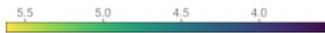

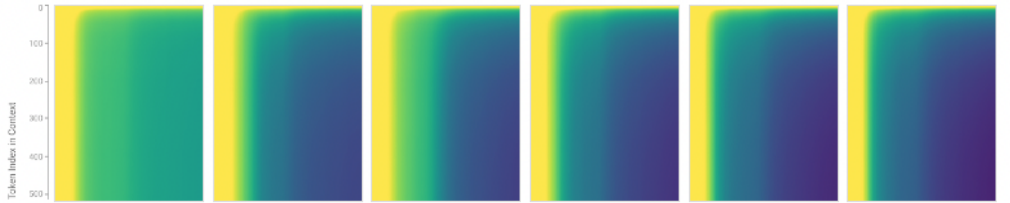

**DERIVATIVE OF LOSS WRT CONTEXT INDEX**

Heatmap of $d\text{Loss}(n_{\text{train}}, i_{\text{ctx}}) \;/\; d\ln(i_{\text{ctx}})$. The log vertical partial derivative of the above graph.
More...

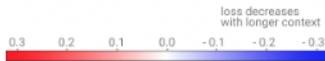

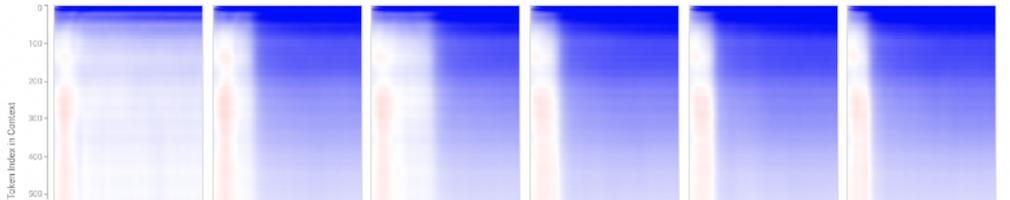

**"IN-CONTEXT LEARNING SCORE"**

$\text{Loss}(n_{\text{train}}, i_{\text{ctx}} = 500)$
$- \text{Loss}(n_{\text{train}}, i_{\text{ctx}} = 50))$

How much better the model is at predicting the 500th token than the 50th? We use this as a proxy for how good the model is at in-context learning. More...

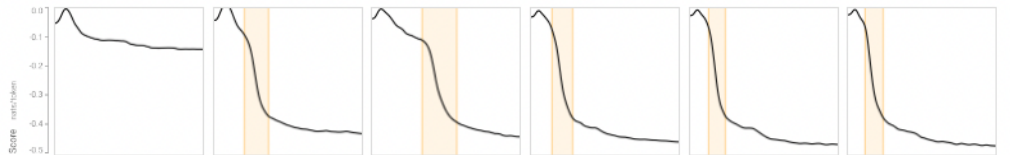

# Model Details

### SMALL MODELS

The small models are 1- through 6-layer Transformers. These include models both with MLPs and without MLPs (i.e. "attention only" models). They have a context window of 8192 tokens, a 216 token vocabulary, an activation dimension dmodel = 768, and 12 attention heads per layer regardless of total model size. They were trained for 10,000 steps (~10 billion tokens), saving 200 snapshots at intervals of every 50 steps. Their positional embeddings are implemented with a variant on the standard positional embeddings (similar to Press *et al*. [17]). The training dataset is described earlier at the start of the Model Analysis Table.

We observe a "phase change" phenomenon that appears at approximately 1-3 billion tokens in the small models. It might be reasonable to ask whether these phenomena are driven by scheduled changes to the hyperparameters, such as learning rate or weight decay. Weight decay was reduced at 4750 steps (approximately 5 billion tokens), the effects of which can be seen as a slight deviation about halfway through the displayed loss curves, occurring at the exact same point for all models; this is not related to the phase change, as this step number is notably beyond the range in which the phase change occurs. The only other hyperparameter change that occurs within the range of the phase change is the learning rate warm-up, which ramps up over the first 1.5e9 tokens.

### FULL-SCALE MODELS

The "full-scale models" are from the same set of models as described in Askell *et al*. [18] The context window and vocabulary size are the same as the small models (that is, 8192 tokens and 216 tokens respectively). Unlike the small models, their dimensions are adjusted to scale up with increasing size, with an activation dimension $d_{model} = 128 * n_{layer}$, and a varying number of attention heads (See Appendix for full details). The models have both dense and local attention heads. In a local attention head, each token may only attend to earlier tokens within a fixed window of relative positions. Dense heads are the standard head, where a token may attend to *any* earlier token (including itself). The training dataset is described earlier at the start of the Model Analysis Table.

Snapshots from these models were saved at exponential step numbers, at an interval of 2x. For our analyses we use snapshots at steps from 2^5 through 2^17, plus one or two final saves thereafter, for a total of 15 saved snapshots (except the 40L which has 14 saved snapshots).[26] This corresponds to a consistent number of tokens across all models up through 2^11 steps (= 2.15E+09 tokens), after which adjustments to the training schedule cause the number of tokens per step to increase for the 24L and 40L models.

TABLE OF MODEL PROPERTIES FOR FULL-SCALE MODELS

| $n_layer$ | Non-embedding parameter counts | Activation dimension $d_{model} = 128 * n_{layer}$ | Attention heads per layer | Attention dimension $d_{head}$ |
|---|---|---|---|---|
| 4 | 13M | 512 | 8 | 64 |
| 6 | 42M | 768 | 12 | 64 |
| 10 | 200M | 1280 | 20 | 64 |
| 16 | 810M | 2048 | 32 | 64 |
| 24 | 2.7B | 3072 | 48 | 64 |
| 40 | 13B | 5120 | 40 | 128 |

SMEARED KEY MODELS

The "smeared key" architecture modification described in Argument 2 is as follows: we introduce a trainable real parameter $\alpha$ used as $\sigma(\alpha) \in [0, 1]$ that interpolates between the key for the current token and previous token:

$$k_j = \sigma(\alpha)k_j + (1 - \sigma(\alpha))k_{j-1}$$

(In the case of the very first token in the context, no interpolation happens). These models were otherwise proportioned and trained exactly the same as the small models. We present these only at one-layer and two-layer sizes.

# Unexplained Curiosities

As with all scientific investigations, in the course of this work we've encountered a few unexplained phenomena. In this section, we discuss these and provide very preliminary investigations of a few that were especially surprising.

## Seemingly Constant In-Context Learning Score

One of the stranger observations in this paper is that in-context learning score (as we've defined it: the loss of the 500th token in the context minus the loss of the 50th token in the context) is more or less the same for all models after the phase change. It appears to not matter whether the model is a tiny two layer model or a fairly large 13 billion parameter model, nor whether the model has just gone through the phase change or trained much longer. The only thing that matters, seemingly, is whether the model has gone through the phase change at all.

A natural question is whether this might be an artifact of the relatively arbitrary definition. Afterall, there's no reason to privilege token index 50 or 500 in the context. But it appears that varying these doesn't matter. In the following plot, we show how the large models' "in-context learning score" varies if we define it instead as the difference between the loss at the final token in the context (8192) and other indices. While there are small differences between models – for some definitions, small models would do slightly more "in-context learning"![27] – all definitions appear to show that the amount of in-context learning varies only slightly between models.

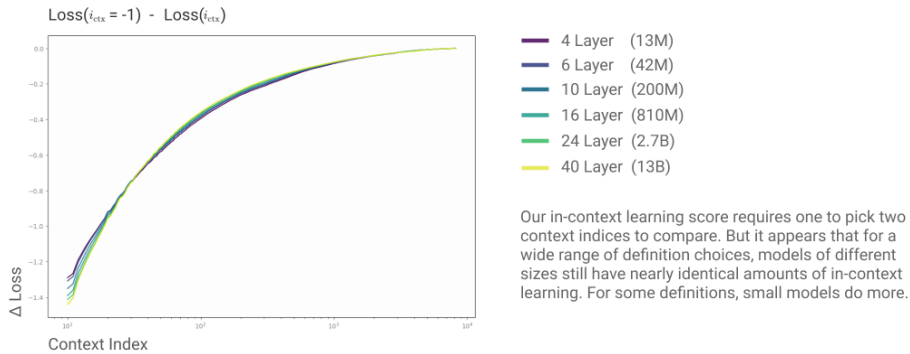

How can this be? First, it's important to be clear that large models still predict tokens at all indices better than small models, and they're best at predicting later tokens. What's going on is that the large models gain all their advantage over small models very early in the context. In fact, the majority of the difference forms in the first ten tokens:

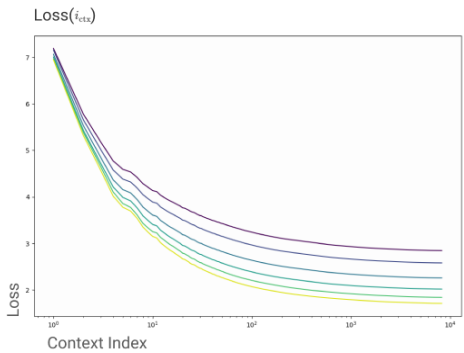
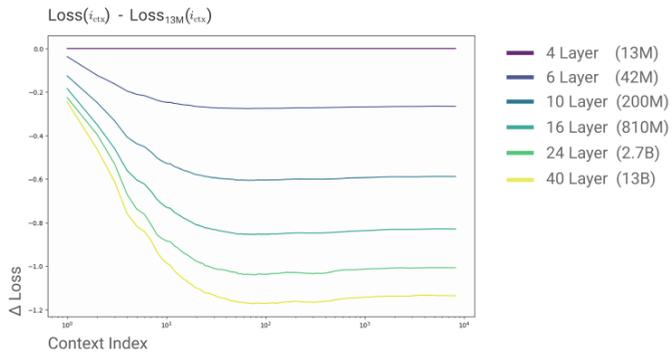

It seems that large models are able to pull a lot of information out of the very early context. (This might partly be, as an example, because their increased world knowledge mean they don't need to gain as much information from the context.) They then further decrease their loss by a roughly fixed amount over the remainder of the context.[28] It seems likely this fixed amount is in some sense "more difficult in-context learning" for large models, since they're starting from a lower loss baseline. While it still seems mysterious to us why models should have the same in-context learning score, this perspective makes it "strange" rather than "shocking".

## Phase Change Effect on Loss Derivatives

Another observation we find quite striking is that if one looks at the derivative of the loss curves of models of different sizes, it appears that their order switches at the phase change. This is most easily seen by plotting the derivative of loss with respect to the log of elapsed tokens (since loss curves are often most easily reasoned about on a log x-axis). The key observation is that the loss decreases more slowly for small models than large models before the phase change, but the opposite is true after.

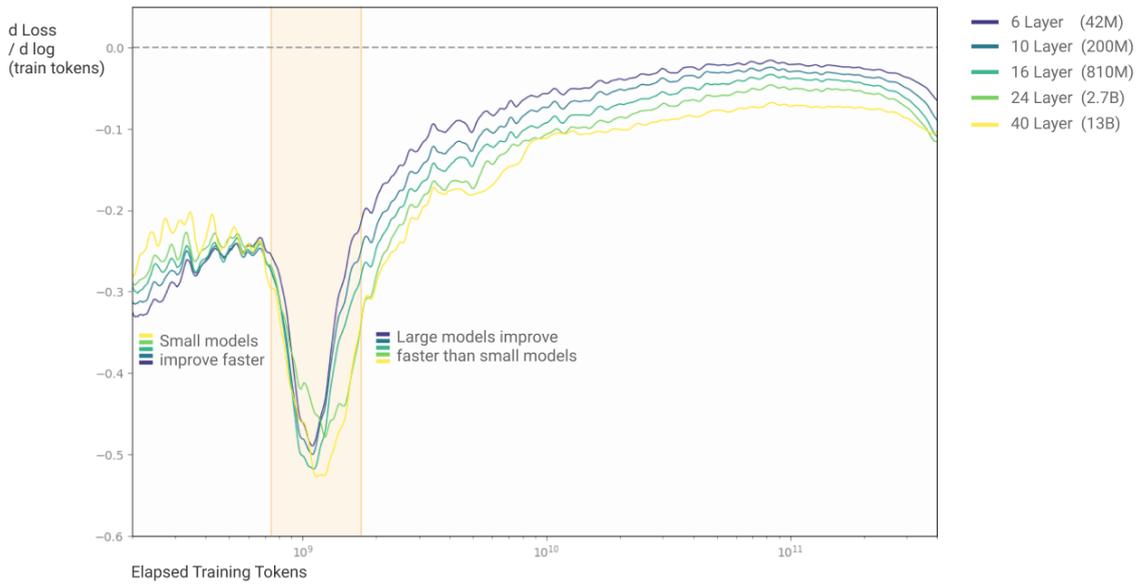

While it doesn't seem that surprising that small models learn more quickly in early training, it is striking that this inversion seems to coincide with the phase change. It's another piece of evidence that suggests the phase change is an important transition point in the training of transformers.

## Additional Curiosities

In the model analysis table:

- The 6-layer attention-only model has an unusual head that develops in the later half of training. This head is *not* an induction head, and yet ablating it has an effect similar to reversing the phase change (in the "before-and-after vector" attribution plot). What is this head?

- The 4-layer MLP model ablations are nowhere near as "peaky" as those of any other model. What is different about this model's development?

- The 6-layer MLP model shows a "loss spike". We don't yet know what causes loss spikes.

- The 6-layer MLP model has one lone induction head whose ablation has the opposite effect on the in-context learning score. What is this head?

And in the Appendix:

- Full-scale models above 16 layers start to show a small number of heads that score well on "prefix search", but get a *negative* score on copying, which means they are not induction heads. What can we learn about these "anti-copying prefix-search" heads?

# Discussion

### Safety Implications

The ultimate motivation of our research is the theory that reverse engineering neural networks might help us be confident in their safety. Our work is only a very preliminary step towards that goal, but it it does begin to approach several safety-relevant issues:

**Phase changes:** If neural network behavior discontinuously changes from one scale to the next, this makes it more challenging for researchers and society to prepare for future problems.

**In-Context Learning:** In-context learning has been a topic of concerned speculation in the safety community. With less-capable neural networks, one might be tempted to treat their behavior as relatively fixed after training. (That said, demonstrations of adversarial reprogramming [21] shed some doubt on this assumption.) In-context learning highlights that model behavior can in some sense "change" during inference, without further training. Even if we think of in-context learning as "locating" an already-learned behavior [22], rather than learning something new, the behavior could be a surprising and unwanted off-distribution generalization.

**Mesa-Optimization:** There have been some concerns that the underlying mechanism of in-context learning might be mesa-optimization [14], a hypothesized situation where models develop an internal optimization algorithm. Our work suggests that the primary mechanism of in-context learning, at least in small models, is induction heads. We did not observe any evidence of mesa-optimizers.

### Linking Learning Dynamics, Scaling Laws, and Mechanistic Interpretability

The in-context-learning phase change may be a useful "Rosetta stone" linking mechanistic interpretability, learning dynamics, [23], and statistical physics-like empirical properties of neural networks (e.g. scaling laws or phase changes). If one wants to investigate the intersections of these lines of work, the phase change seems like an ideal starting point: a concrete example where these lines of inquiry are intertwined, which can be explored in small models, bounded in a small sliver of the training process, and is linked to a capability (in-context learning) the community is excited about.

# Related Work

The general approach of this paper to reverse engineering transformers is based heavily on our previous paper, _A Mathematical Framework for Transformer Circuits_. There is much to be said about how that framework relates to other work in interpretability. Rather than repeating it, we refer readers to Related Work in our previous paper, especially discussion of the relationship to circuits [6], to analysis of attention heads (e.g. [24, 25, 26, 27, 28]), and to related mathematical analysis (e.g. [29]).

Building on that perspective, we here focus on how aspects of this paper raise new connections to the machine learning literature, separate from the connections raised simply by the underlying framework.

### IN-CONTEXT LEARNING

Emergent in-context learning was compellingly demonstrated in GPT-3 [1]. A number of papers have studied how to effectively leverage in-context learning, especially with "prompt engineering" [11, 12]. But of particular importance to us, several papers have tried to study how and when in-context learning occurs (*e.g.* [13, 30, 31, 32]).

Some of the findings of these papers are consistent with the induction head hypothesis, or support our methodology:

- Kaplan *et al*. [13] is the origin of our approach for using loss at different token indices as a formalism for studying in-context learning.
- O'Connor & Andreas [30] find that preserving word order in contexts is important, as the induction head hypothesis would suggest.

However, there are also places where experiments in these papers seem in tension with the induction head hypothesis:

- O'Connor & Andreas [30] have some experiments suggesting that removing all words except nouns can improve loss. This seems inconsistent with the induction head hypothesis. However, they only find this for experiments where they retrain the model on modified data. This seems both less directly related to our work (because we aim to study models trained on natural data) and subtle to interpret (because retraining models introduces the possibility of run-to-run loss variation, and the measured loss differences are small). The experiments where they don't retrain models on modified data seem consistent with the induction head hypothesis.
- Xie *et al*. [31] finds that LSTMs outperform Transformers when fit to synthetic data generated by a Hidden Markov Model (HHM) designed to isolate a particular theoretical model of in-context learning. We generally expect Transformers to outperform LSTMs at in-context learning on natural text (as seen in Kaplan *et al*. [13]), with induction heads as a major explanation. But in the case of the Xie *et al*. experiments (which don't use natural text), we suspect that the structure of the synthetic data doesn't benefit from Transformers, and that LSTMs are perhaps better at simulating HMMs.

Note that we are using a broader conception of "in-context learning", rather than something as specific as "few-shot learning". This is in contrast with Brown *et al*. [1], which describes that a language model "develops a broad set of skills and pattern recognition abilities. It then uses these abilities at inference time to rapidly adapt to or recognize the desired task," with examples of tasks such as few-digit addition and typo correction. In our conception of "in-context learning", we refer to all the ways that a model rapidly adapts to or recognizes what is going on in the context, even if "what is going on in the context" isn't well-conceived-of as multiple "shots" of some other specific repeated task.

### SCALING LAWS

Over the last few years, the observation that machine learning models change in smooth, predictable ways described by scaling laws [13] has emerged as a useful tool for modelling the properties of models before training them.

The relationship between scaling laws and mechanistic interpretability might be seen as analogous to the relationship between thermodynamics and the physics of individual particles. For both thermodynamics and scaling laws, even though the underlying system is very complicated, we're able to find simple relationships between the variables – for thermodynamics: entropy, temperature, volume and pressure; for neural networks: loss, compute, parameters, and data. In contrast, mechanistic interpretability studies the individual circuits underlying our models, vaguely analogous to how one might carefully study individual particles in physics. In physics, these two layers of abstraction were bridged by statistical physics.

Can we bridge these two levels of abstraction in machine learning? The induction head phase change is the first time we're aware of a bridge. They give us a phenomenon at the level of macroscopic properties at loss which can be explained at the level of circuits and mechanistic interpretability.

In fact, induction heads may be able to explain previous observed exceptions to scaling laws. In our work, 1-layer transformers seem very different from deeper transformers. This was previously observed by Kaplan *et al*. [13] who found that 1-layer transformers do not follow the same scaling laws as larger transformers. It seems quite plausible that the reason why the scaling laws are different for one-layer models is that they don't have induction heads.

**PHASE CHANGES & DISCONTINUOUS MODEL BEHAVIOR**

In the previous section, we discussed how scaling laws describe smooth predictable relationships between model loss and properties like scale. However, a more recent set of results have made the situation seem more subtle. While models' losses often scale in predictable ways, there are cases where behavior is more complex:

- Brown *et al*. [1] find that, while aggregate loss scales predictably with model size, models' ability to perform specific tasks like arithmetic can change abruptly.

- Power *et al*. [8] observe a phenomenon they call "grokking" where models discontinuously jump from random chance to perfect generalization as they train.

- Double Descent [33] is a phenomenon where model performance first gets worse due to overfitting as one makes a model larger (the "classical" regime), but then gets better again past a certain point (the "modern" regime). Generalizations of double descent can occur with respect to parameter size, dataset size, or amount of training [34]. These phenomena are not discontinuous in loss, but they are surprising trend reversals, and perhaps discontinuous in derivatives.

For more general discussion of these phase change phenomena, see a recent blog post by Steinhardt [35].

The discontinuous phase change behavior we observe with induction heads is most analogous to Power *et al*. [8]'s "grokking", in that it occurs over the course of training. We think our main contribution to this literature is linking the changes we observe to the formation of induction heads and a parameter-level understanding of the circuits involved. As far as we know, induction heads are the first case where a mechanistic account has been provided for a phase change in machine learning.

LEARNING DYNAMICS

If neural networks can genuinely be understood mechanistically, in terms of circuits, it seems like there almost *has to* be some way to understand the learning process in terms of the dynamics of circuits changing. Induction heads offer an interesting preliminary bridge between these topics, and are a source of optimism for such a connection. This section will briefly review some strands of work on the learning dynamics side, which seem particularly promising to think about if one wanted to pursue such a connection further.

One remarkable result in learning dynamics, by Saxe *et al* [23], has been the discovery of *closed form* solutions to learning dynamics for linear neural networks without activation functions. The exciting thing about this work is that it actually provides a simple way to conceptually think about neural network learning in a simplified case. (In follow up work, Saxe *et al* also explore connections between this framework and models learning to represent semantic information [36].) We're unable to provide a detailed review of this work here, but we note that Saxe *et al*'s framework could naturally suggest a circuit lens for thinking about learning dynamics. Very roughly, they find that linear neural networks can be understood in terms of the evolution of independent paths through the network, with each path corresponding to a principal component of the data. These paths might be thought of as circuits.

Another interesting line of work has been the study of the geometry of neural network loss surfaces (*e.g.* [37, 38, 39, 40]). Here, our thoughts on the connection are more superficial, but it seems like there must be some way in which aspects of the loss surface connect to the formation of circuits. Very concretely, it seems like the phase change we've described in this paper must correspond to some very large feature in the loss landscape of transformers.

UNIVERSALITY

In the context of interpretability and circuits, "universality" [41] or "convergent learning" [42] is when multiple models develop the same features and circuits. Universality might seem like an intellectual curiosity, but the circuits thread argues that universality plays a critical role in what kind of interpretability makes sense:

> [I]magine the study of anatomy in a world where every species of animal had a completely unrelated anatomy: would we seriously study anything other than humans and a couple domestic animals? In the same way, the universality hypothesis determines what form of circuits research makes sense. If it was true in the strongest sense, one could imagine a kind of "periodic table of visual features" which we observe and catalogue across models. On the other hand, if it was mostly false, we would need to focus on a handful of models of particular societal importance and hope they stop changing every year. [41]

Research on universality began with *Li et al.* [42] who showed that many neurons are highly correlated with neurons in retrained versions of the same model. More recently, a number of papers have shown that in aggregate, neural networks develop representations with a lot of shared information (*e.g.* [43, 44]). The Circuits thread tried to extend this notion of universality from features to *circuits*, finding that not only do at least some families of well-characterized neurons reoccur across multiple networks of different architectures and that the same circuits [41], but the same circuits appear to implement them [45].

Certain kinds of universality are often implicitly assumed in the language model attention head interpretability literature. For example, it seems widely accepted that "previous token" attention heads form across many transformer language models (*e.g.* [26, 27]). The implicit hypothesis of universal attention heads – that is, attention heads with the same attention patterns in different models – isn't exactly the same thing as the kind of feature universality studied in the vision context, but is kind of analogous.

Our work in this paper has analogies to many of these strands of prior work. Like the previous attention head papers, we describe the induction head pattern as a universal attention pattern. However, our analysis of these heads' OV and QK circuits extends this claim of universality to the circuit level, similar to the original Circuits thread. And a corollary of our analysis of the OV circuit is a claim about what feature the attention head computes (roughly: the token embedding of the token following a previous copy of the present token) which is more similar to the traditional work on universality.

Separate from all of this, it's worth mentioning that increasingly there's evidence for a particularly extreme kind of universality at the intersection of neuroscience and deep learning. Increasingly, research suggests that biological and artificial neural networks learn similar representations (*e.g.* [46, 47, 48]). In fact, Goh *et al*. [49] find that multimodal "concept" neurons found in humans (such as the famous "Jennifer Anniston neuron") occur in neural networks.

ATTENTION PATTERNS IN TRANSLATION-LIKE TASKS

In Argument 4, we saw an induction head that helps implement translation. Although we're not aware of anything quite so general in the prior literature, there are reports of attention patterns which, in retrospect, seem somewhat similar. Often, in translation-like tasks, we see attention attend to the token which is *about to be translated*. We see this in literal translation (*e.g.* [50]) and also in voice recognition (*e.g.* [51] where the model attends to the portion of the audio about to be transcribed). Visualizations of this in the encoder-decoder context often slightly obscure the induction-like nature of the attention patterns, because the decoder is visualized in terms of the output tokens predicted per time step rather than its input tokens.

# Comments & Replications

*Inspired by the original Circuits Thread and Distill's Discussion Article experiment, the authors invited several external researchers who were also investigating induction heads to comment on this work. Their comments are included below.*

## REPLICATION

*[Adam Scherlis](#) is a researcher at [Redwood Research](#).*

Redwood Research has been working on language model interpretability, inspired in part by Anthropic's work. We've found induction heads reliably appearing in two-layer attention-only transformers. Their structure roughly matches the description in ["Analyzing a Two-Layer Model"](#) from Anthropic's previous paper, with previous-token heads in layer 0 and induction heads in layer 1 (often one of each). These each have the expected attention behaviors. We tested this by replacing attention-score matrices with idealized versions and comparing the change in loss to the change from ablating that head. Replacing the previous-token head's scores with exact previous-token attention recovered 99% of the loss difference. Replacing the induction head's scores with a simple approximation (attending to the first token and to exact `[A][B]...[A]` matches) recovered about 65% of the loss difference. Our induction heads also match patterns of the form `[A][B][C]...[A][B]→[C]`; including this in the substituted attention scores recovered an additional 10% of loss difference. The induction head's OV circuit copies tokens, including some fuzzy matching of semantically similar tokens. Its QK circuit is dominated by K-composition with the previous-token head; the previous-token head's OV matrix copies information to a new subspace and the induction head's QK matrix copies it back to the usual token embedding. We've also cataloged a few other kinds of attention heads, including skip-trigram heads.

## REPLICATION

*[Tom Lieberum](#) is a master's student at the University of Amsterdam.*

I used the empirical criterion for induction heads [presented](#) in this paper to look for induction heads in publicly available models. To reiterate: on a sequence of tokens `[A][B] .... [A] → [B]`, a head is called an induction head if it attends to the previous `[B]` when reading the last `[A]`, and if the previous `[B]` increased the logit of the last `[B]`.

Under this definition, I found potential induction heads in GPT2 and GPT-Neo mostly starting in the mid-depth region. I made an interactive version to explore attention and logit attribution across all layers and heads on the long Harry Potter prompt with the repetition of the first paragraph. It can be accessed [here](#). For example, for GPT2-XL, head 20 in layer 21 seems to be an induction head, as well as head 0 in layer 12 of GPT-Neo-2.7B. For these heads we can see that virtually every token in the repetition of the first paragraph attends to its following token in the original paragraph. Thanks to [EleutherAI](#) for providing the compute resources for this project.

At one point the paper speculates on the minimal context length necessary to form induction heads. On a synthetic dataset, I have already found induction to be easily learned with a context length of 4. However, I believe that this is, in general, a property of the dataset/data distribution rather than the model itself, i.e. "how useful is induction for this task, on this dataset?". While it is intriguing to think about the statistics of language in this way, I am not sure how useful this line of research would be to the overall project of interpretability.

## Change Log

The following changes have been made after publication:

- 2022-09-20: Corrects the definition of Prefix matching in two places, which incorrectly stated "tokens that were *preceded* by the current token" instead of "preceded" or "were *followed* by". Adds detail to the definition of the copying head evaluator to explain that the effect on "raising the logits" is calculated using a ReLU.

## Footnotes

1. Note that *mechanistic interpretability* is a subset of the broader field of *interpretability*, which encompasses many different methods for explaining the outputs of a neural network. Mechanistic interpretability is distinguished by a specific focus on trying to systematically characterize the internal circuitry of a neural net. [↩]

2. Note that induction heads don't occur in 1 layer models, because they require a composition of attention heads in different layers. [↩]

3. Or sometimes *metalearning*, although this term implicitly makes the stronger claim that the model is learning *how* to do a new ability (as opposed to "locating" an ability, i.e. learning *what* it is supposed to do), an implication which is both controversial and not entirely precise in its meaning. [↩]

4. More specifically, induction heads seem to largely decouple `A` and `B`. While some induction heads may specialize on certain kinds of `A` s or `B` s, this significant decoupling of `A` and `B` means that they don't have a fixed table of bigram statistics they can update on, but rather can abstract to new patterns. [↩]

5. In practice, induction heads don't exhibit these properties perfectly, and our measurements give us a continuum, but there is a clear subset of heads which exhibit these properties with much greater than random chance. [↩]

6. By defining induction heads in terms of their behavior on repeated copies of random sequences, we can be confident that it's actually relying on induction rather than, say, a simple copying head that heuristically attends to previous tokens which could plausibly slot in well after the next token, even if that hasn't occurred yet in the current context. [↩]

7. The simplest induction heads match just one preceding token. But we also often observe induction heads that perform a fuzzy match over several preceding tokens. [↩]

8. In mathematics, one sometimes thinks of a function as an infinite dimensional vector of the values it would give for different inputs. For neural networks, this can be a nice way to abstract away the fact that functionally identical models can have very different parameter vectors. Of course, we can't directly represent these infinite dimensional vectors, but we can approximate them by sampling. [↩]

9. Although see later discussion for a way in which a constant 0.4 nats can be viewed as a "larger" improvement for a more-powerful model, because attaining the same magnitude of improvement starting from a better baseline is more challenging. [↩]

10. As we'll see, many other curious phenomena occur during the phase change as well. [↩]

11. That said, it's also the case that aside from the experiments varying the dataset, all these models were trained on the same tokens in the same order [↩]

12. Of course, this passage is merely a single piece of anecdata, shown here to provide a qualitative intuition rather than meant as systematic evidence. For a more comprehensive analysis of the observed behavioral change before and after the phase change, consult the "ablation to before-and-after vector" analyses in the Model Analysis Table [↩]

13. Some bits of information are harder to learn than others. For example, near the start of training, the model can achieve a significant decrease in loss with simple approaches such as memorizing bigram statistics. But later in training all the low-hanging fruit has been plucked, and the model must learn more sophisticated algorithms such as induction heads, which is likely harder to learn yet results in fewer bits of information. In today's state of the art models, such as GPT-3, we observe complex capabilities such as addition, and we can imagine future models approaching near-perfect loss might need human-level understanding of language and beyond to get those last few fractions of a bit. [16]  So the *relative* loss between token 50 and token 500 likely represents harder and harder bits over the course of training. [↩]

14. It is still a mystery why these forces of increased capacity for in-context learning and increasing difficulty of marginal bits would so *exactly* balance, and this seems a promising avenue of future research. Perhaps there is not, in fact, much

14. overlap in the information that could be gained from both in-context learning and other approaches, so *these* bits do not get harder with time. We'll revisit this in the Discussion. [↩]

15. Why do induction heads require composition? In an induction head, where the head attends must be a function of the token before the token it attends to. But an individual attention head computes attention scores only from the source and destination token. Without using information written by a second earlier attention head, the score from the attended token can't be a function of the token that precedes it. [↩]

16. The key vector for the first token is unchanged. [↩]

17. In fact, ablating most other attention heads appears to *increase* in-context learning. At first, that seems kind of crazy: how could damaging the model make it *better* at in-context learning? It appears that some tokens can be predicted with both "normal prediction" and in-context learning. If ablating a head makes the model bad at "normal prediction", in-context learning can predict more tokens that otherwise wouldn't be predicted, and in-context learning score as we've defined it increases. [↩]

18. Note that the cost of a full set of ablations scales superlinearly with model size at $O(N^{1.33})$, since there are $O(N^{0.33})$ heads and each ablation is $O(N)$ where $N$ is the number of parameters. The base cost of an ablation is also non-trivial, since we're evaluating each ablation on 10,000 examples, for each training checkpoint. [↩]

19. In attention-only models, the logits can be expressed (up to a rescaling due to LayerNorm) as a sum of terms from each attention head, along with a "direct path term" to the the token embedding (see our previous paper). The direct path term is solely a function of the present token, so it can't contribute to in-context learning. That means that all in-context learning must ultimately originate with attention heads, and since the relationship is almost linear, ablations (with frozen attention patterns) are a principled way to measure their contribution. [↩]

20. Why is ablating attention heads harder to reason about if they can interact with MLP layers? At a high-level, the issue is that in-context learning is a complicated function of which heads are ablated, rather than a sum of their contributions. But it may be helpful to consider specific examples. One possibility is that ablating a head might shift the statistics of an MLP layer and "break" neurons by shifting their effective bias, without actually having a meaningful role. Another possibility is that an important MLP layer mechanism relies on two attention heads, but can function reasonably well with one if the other is ablated. [↩]

21. We compute this by taking the value vector produced at each position, weighting it by the attention matrix, and multiplying it by $W_O$ and the unembedding, and selecting the logit value for the corresponding token. Note that we first normalise the vector of logits to have zero mean, as adding a constant to every argument in a softmax has no effect. [↩]

22. For example, a special case of translating from English to another language is translating English to itself, which is precisely the same as literal copying. [↩]

23. The only other potential contender for driving in-context learning in two-layer attention only models would be basic copying heads. However, basic copying heads also exist in one-layer models, which don't have the greatly increased in-context learning we see in two-layer models. Further, induction heads just seem conceptually more powerful. [↩]

24. Note that the attended token is only ignored when calculating the *attention pattern* through the QK-circuit. It is extremely important for calculating the head's *output* through the OV-circuit! As observed in our previous work, the parts of the head that calculate the attention pattern, and the output if attended to, are separable and are often useful to consider independently. [↩]

25. Instead, they use a slightly unusual positional mechanism similar to Press *et al*. [17] [↩]

26. Note that this is different to the small models, which have 200 snapshots saved at linear intervals. As the full-scale models have only 14 or 15 snapshots, this makes it harder to judge the shape of the curves as confidently as for small models. [↩]

27. Around token 100, there's a regime where small models reduce their loss very slightly more per token than large models. We interpret this as the small models picking up low-hanging fruit that the large models already got in the very early context. [↩]

28. In fact, small models gain slightly more in the mid-context, catching up a tiny bit with large models, but it's a small effect. [↩]

29. The fact that this is possible at all is a surprising application of the insight in our previous work that the calculation of each attention head's attention pattern and output vector is *separable*, that is, implemented by different circuits consisting of different sets of parameters which may be run and studied in isolation [↩]

30. As an interesting aside, we note that with the attention pattern frozen the action of each head is purely linear, which means that in an attention-only model, a pattern-preserving ablation is equivalent to subtracting all terms containing the ablated head from the output logits. See our previous work for further exploration of the implications of this. This does not hold in more complex models, due to the non-linearity introduced by MLPs. [↩]

31. It would've been better if we had used a start of sequence token for the copying head evaluator as well, but we omitted it by mistake. Without the "start of sequence" token, some heads that were doing prefix matching on real data would get anomalously low scores on our test sequences [↵]

## Acknowledgments

In writing this paper, our thinking and exposition was greatly clarified by detailed correspondence with Sam Bowman, Paul Christiano, Aidan Gomez, Dan Hendrycks, Jacob Hilton, Evan Hubinger, Andrew Ilyas, Percy Liang, Tom Lieberum, Chris Maddison, Aleksander Madry, Ethan Perez, Jacob Steinhardt, and Martin Wattenberg.

We're especially indebted to Adam Scherlis and Tom Lieberum who replicated our results and wrote comments to be included with this paper.

We're also deeply grateful to Daniela Amodei, Stanislav Fořt, Tristan Hume, Saurav Kadavath, Jamie Kerr, Shauna Kravec, Jeffrey Ladish, Jia Yuan Loke, Liane Lovitt, Rebecca Raible, Sheer El Showk, Timothy Telleen-Lawton, Matt O'Brien, Kate Rudolph, Jemima Jones, Geoffrey Irving, Tom McGrath, Michela Paganini, Allan Dafoe, Gabriel Goh, Nick Cammarata, Chelsea Voss, Shan Carter, Katherine Lee, Beth Barnes, Jan Leike, Nate Thomas, Buck Shlegeris, Alex Tamkin, Quinn Tucker, and Rob Harries; for their support, for comments on this work, and for conversations that contributed to the background thinking on interpretability and safety this work is based on.

## Author contributions

**Research**: The bulk of this line of work, including many iterations of experiments that didn't make it into the final paper, was conducted by Catherine Olsson with mentorship from Chris Olah. Chris first discovered the "induction bump" in a 2-layer model. He had the idea of investigating what aspects of the per-token loss changed most over the bump, and made the connection to in-context learning. Catherine led data collection and analysis throughout the project. Data collection made extensive use of Garcon infrastructure by Nelson Elhage. Nelson collected data and ran analyses for the smeared key models. Chris Olah, Neel Nanda, and Catherine Olsson wrote the head activation evaluators. Research ideas were significantly influenced and improved by others at Anthropic.

**Writing:** This article was drafted by Catherine Olsson, Chris Olah, and Nelson Elhage, with contributions and editing from Neel Nanda and Dario Amodei. Dario Amodei originally suggested the "arguments" structure. The specific arguments were iteratively refined in collaborative conversations between Dario Amodei, Chris Olah, Catherine Olsson, and Nelson Elhage. Nick Joseph contributed significant editing. Other members of Anthropic made miscellaneous contributions throughout the writing process.

**Figures and visualizations:** Catherine Olsson generated the graphs in the model analysis table, and did preliminary design and styling. Chris Olah finalized the design and styling, and created "pull-out" versions for smaller figures earlier in the paper. Nelson Elhage contributed to the graphs of per-token losses on Harry Potter. Chris discovered the heads showcased in Claim 2 and designed the "month-animal-color-fruit" task. Nelson set up the interactive visualizations, using existing visualization tools built by Chris, with contributions from Nelson and Catherine, using PySvelte (see github).

**Model training:** Nelson Elhage created the attention-only models and the smeared-key models. Dario Amodei collected data for models trained on the alternate books-only dataset. All our interpretability research is enabled by having access to models to study, including large models. Led by Tom Brown, Sam McCandlish, and Jared Kaplan, the majority of Anthropic's technical staff contributed to the development of our efficient distributed training infrastructure and the underlying machine learning. Core contributors include Nicholas Joseph, Tom Henighan, and Ben Mann. Nelson Elhage, Kamal Ndousse, Andy Jones, Zac Hatfield-Dodds, and Danny Hernandez also contributed to this infrastructure.

**Cluster:** Tom Henighan and Nova DasSarma, advised by Tom Brown and Sam McCandlish with contributions from many others at Anthropic, managed the research cluster our research depended on and maintained its stability. Nova provided important support when interpretability required unorthodox jobs to run on our cluster.

## Citation Information

Please cite as:

```
Olsson, et al., "In-context Learning and Induction Heads", Transformer Circuits Thread, 2022.
```

BibTeX Citation:

```
@article{olsson2022context,
   title={In-context Learning and Induction Heads},
   author={Olsson, Catherine and Elhage, Nelson and Nanda, Neel and Joseph, Nicholas and DasSarma, Nova and Henighan, Tom and Mann, Ben and Askell, Amanda and Bai, Yuntao and Chen, Anna and Conerly, Tom and Drain, Dawn and Ganguli, Deep and Hatfield-Dodds, Zac and Hernandez, Danny and Johnston, Scott and Jones, Andy and Kernion, Jackson and Lovitt, Liane and Ndousse, Kamal and Amodei, Dario and Brown, Tom and Clark, Jack and Kaplan, Jared and McCandlish, Sam and Olah, Chris},
   year={2022},
   journal={Transformer Circuits Thread},
   note={https://transformer-circuits.pub/2022/in-context-learning-and-induction-heads/index.html}
}
```

## Where induction heads form

In this section we examine: Where are induction heads located in models? Do they generally form in the second layer, or the last layer, or somewhere else?

**Small attention-only models**

In our attention-only models, induction heads form in the last layer.

(In these plots, each point represents a head, colored by layer and arranged along the x-axis by depth in the model. The height of the point is the prefix matching score of that head.)

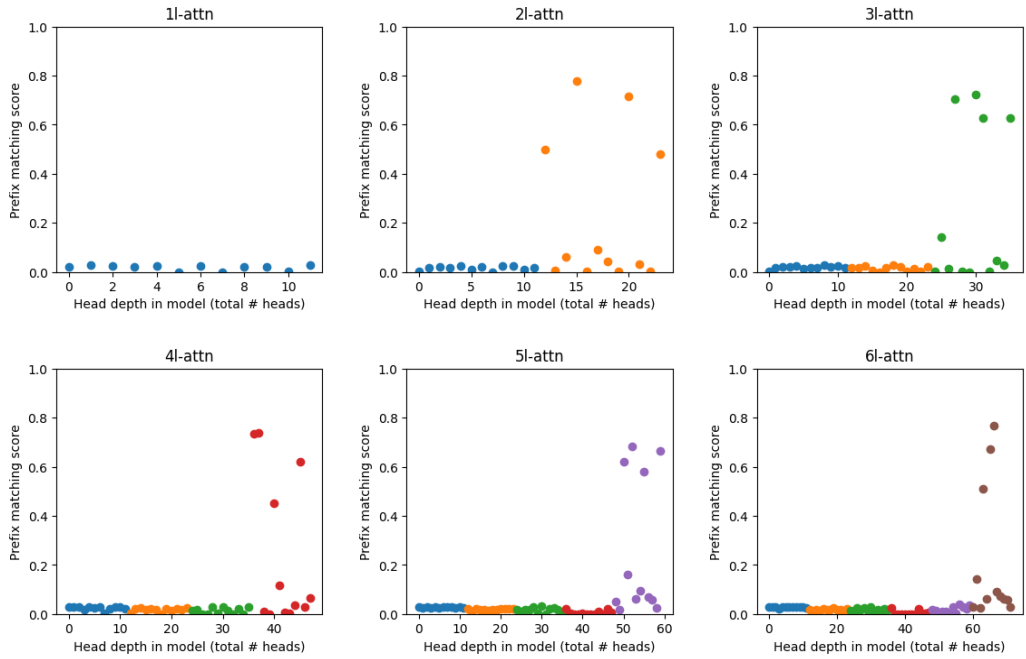

**Small models with MLPs:**

In small models with MLPs, by the 5L and 6L size, the induction heads form more in the *second-to-last* layer.

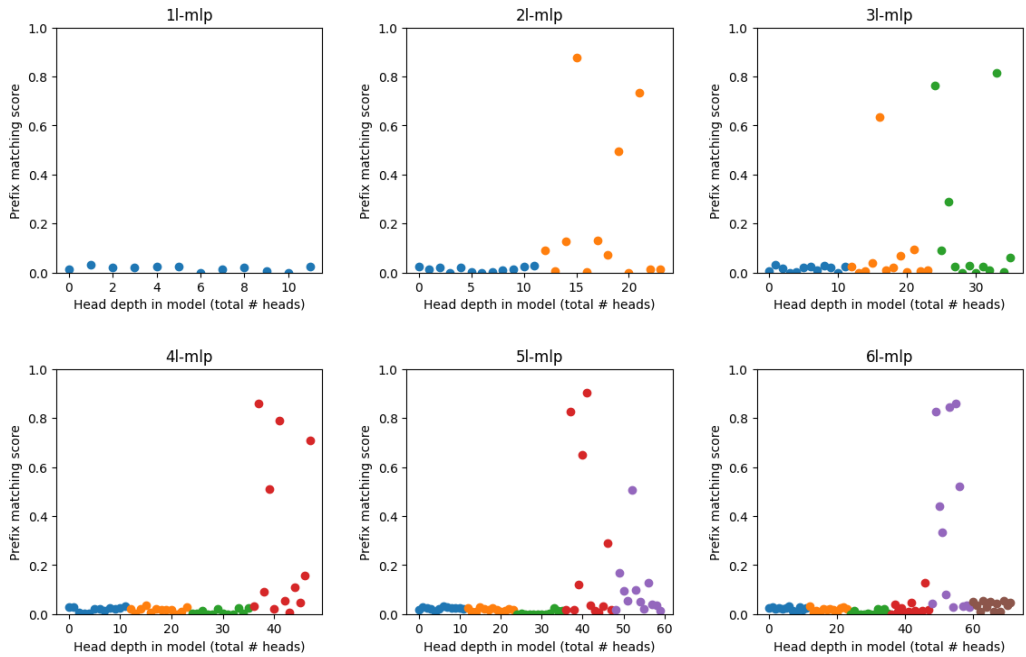

**Full-scale models:**

In the models in our full-scale sweep, induction heads form earlier. In fact, in 24L and 40L models the majority of induction heads form before the halfway point in model depth.

Note that in these plots, only a randomized sample of 100 heads (with selection biased towards those scoring highly on prefix matching) are shown as opaque dots; most of the datapoints are left translucent for ease of reading.

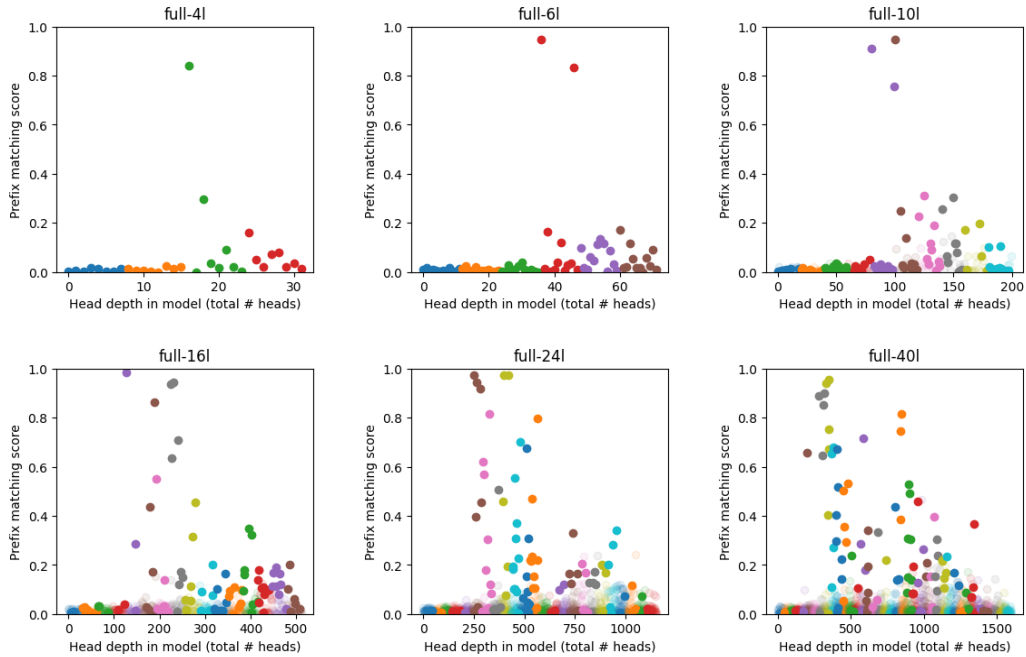

### Distribution of scores

What is the distribution of heads' scores on our head activation evaluators?

The distribution is sharper in our small models than our full-size models. In both cases, heads with a positive prefix matching score are more likely to have a positive copying score, and this correlation is even stronger for the highest prefix matching scores.

(For models with more than 100 heads, a subset of heads is plotted, as seen in the opaque versus translucent datapoints immediately above.)

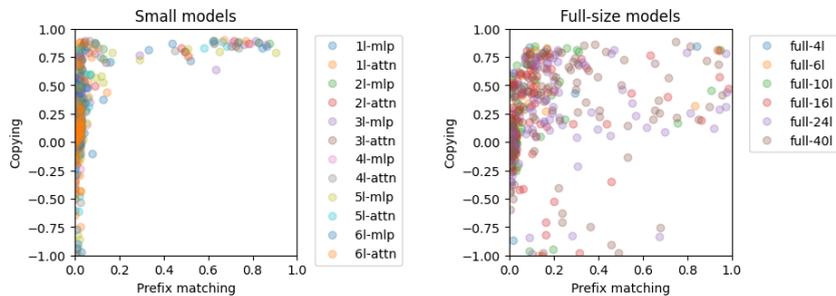

### Validating head activation evaluators

We show here that our head activation evaluators (that we use to measure the properties in our definition of induction heads) correlate well with mathematically-based measures in small attention-only models.

**Copying**

Our copying evaluator correlates well with the OV matrix of each head. Specifically here we plot against the sum of the eigenvalues, divided by the absolute value of the eigenvalues, which is a measure of what fraction of the eigenvalues are positive vs. negative.

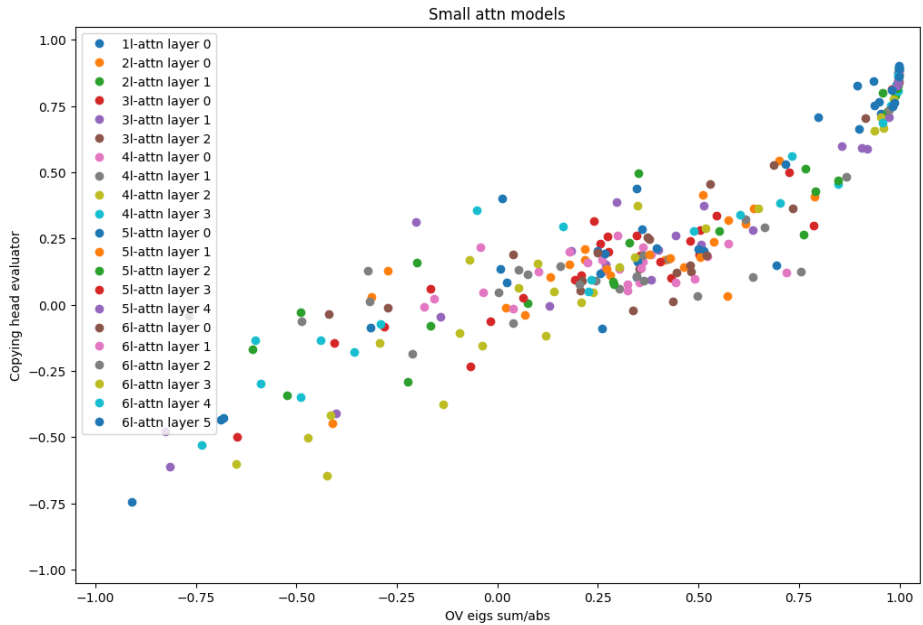

**Prefix matching**

Our prefix matching evaluator correlates well with the trace of QK eigenvalues for the previous-token QK-circuit term.

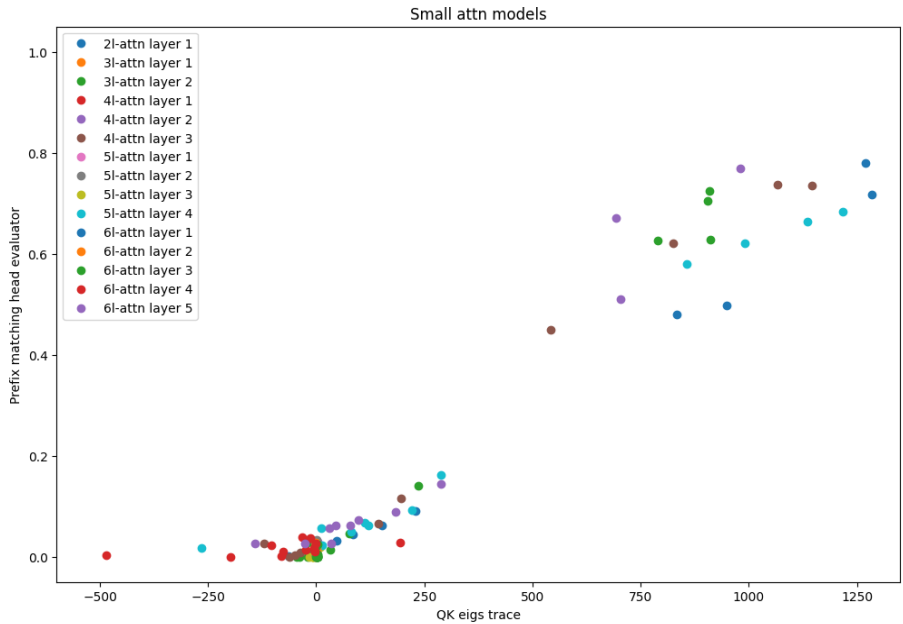

## Data Collection

We measure several properties of the model's overall behavior and of each constituent attention head. With the exception of the attention head ablations (which are the most computationally-intensive, and so were run only on the small models), we measure these properties on all snapshots of all models across training (that is, 200 snapshots each for the twelve small models, and 14 or 15 snapshots each for the six full-scale models). By measuring them on each snapshot, we can see how each property evolves over the course of training, rather than simply measuring the final model.

The properties we collect data on are as follows:

1. **Per-token losses** with **unaltered models**

2. **Per-token losses** with **ablations** of individual attention heads

3. **Empirical head measurements** with **head activation evaluators:** using heuristics to score attention heads based on their observed activations, as another way of estimating "copying", "prefix matching" and "previous token" attention.

4. **Mathematical head measurements** with the **trace of QK eigenvalues** for the previous-token QK-circuit term: measuring how much the WQWK weight matrix reads in a subspace affected by heads that attend to the previous token.

**(1) Per-token losses (unaltered models)**

The most straightforward property we measure is per-token losses from each model snapshot. We run the model snapshot on a set of 10,000 randomly-selected examples from the dataset, each of which is 512 tokens in length, and save individual token losses for every token. The set is held the same between all snapshots and between all models in all three families.

When analyzing the per-token loss data, we summarize it in two different ways:

- **Random token per example.** For each example, we extract the loss for just one random token out of the 512 tokens and discard the rest, giving a length-10,000 vector for each model snapshot. (The random index used for each example is consistent between each model.)

- **Context index average.** We average the per-token losses by context index (i.e. by position), giving a length-512 vector for each model snapshot. The $i_{context}$th entry in the vector is an average over the 10,000 tokens drawn from position $i_{context}$ in each example.

**(2) Per-token losses (attention head ablations)**

One way to study individual attention heads is to measure the difference between the complete model's overall behavior and the overall behavior of a model that is exactly the same except for missing that head. We do this by successively ablating (i.e. zeroing-out) each attention head's contribution to the residual stream. The approach of removing a head's contribution can be thought of as loosely similar to a genetic "knockout" experiment in biology, where observing the effect of removing a specific gene can provide information about the gene's function.

We could do these ablations in two different ways: a "full" ablation, and a "pattern-preserving" ablation. (The [model analysis table](#) shows only pattern-preserving ablations; we explain the other method for comparison) The full ablation would remove *all* of a head's contributions to model behavior, whereas the pattern-preserving ablation removes the ablated head's contribution to all later heads' *output vectors*, but preserves the ablated head's contribution to all later heads' attention *patterns*.[29]

**Full ablation**

- *Method*: Run the model, replacing the ablated attention head's result vector with a zero vector.

- *Outcome*: All downstream computations are impacted (including all Q, K, and V calculations, and attention patterns).

**Pattern-preserving ablation (shown in [Model Analysis Table](#))**

- *Method:*

    - Step 1: Run the model for the first time, saving all attention patterns and discarding the output logits.

    - Step 2: Run model a second time. Replace the ablated attention head's result vector with a zero vector, and force all attention patterns to be the version recorded in the first run.

- *Outcome*: Only downstream V calculations are affected by the ablation, not Q and K calculations. The contribution a head makes to later layers' attention patterns is preserved.[30]

This is by far the most computationally-intensive measurement we perform, although fortunately it consists of many small single-GPU jobs and so we could often run it in "spare" capacity.

As a quick note of comparison, our previous work in Transformer Circuits also contained ablation experiments. Although there are similarities (most notably the decision to freeze patterns), the specific ablations we describe here are *individual attention head* ablations where only one attention head is modified at a time, whereas the ablations described in our prior work are **nth-order term** ablations in which all attention heads are manipulated at once and the process is repeated iteratively *n* times.

**(3) Head activation evaluators**

We construct three heuristic measurements of attention heads, to score how strongly a head exhibits specific properties of interest. These are empirical measurements, based on observed activations on example data.

We evaluate heads on the following properties:

- **Copying.** Does the head's direct effect on the residual stream increase the logits of the same token as the one being attended to?

- **Prefix matching.** On repeated sequences of random tokens, does the head attend to earlier tokens that are followed by a token that matches the present token?

- **Previous token attention**. Does the head attend to the token that immediately precedes the present token?

(Note that "copying" is a property that depends on a head's OV circuit, whereas "prefix matching" and "previous token attention" are properties that depend on a head's QK circuit.)

We are interested in these properties in particular — copying, prefix matching, and previous token attention — because they are relevant to induction heads' function of predicting repeated sequences of random tokens. Specifically, (1) a head with both a "copying" OV circuit and a "prefix matching" QK circuit is thereby an induction head by our definition, because it attends to previous instances of the present token and increases the logit of the token that followed; and (2) a "previous token" head can be used as an algorithmic component of a future induction head in a later layer (as we explain in more detail in Argument 5).

In more detail, each head evaluator is implemented as follows, and each score is averaged over 10 examples.

- *Copying*: Generate a sequence of 25 random tokens, excluding the most common and the least common tokens. Compute this head's contribution to the residual stream, then convert that using the unembeddings (i.e. along the "direct path") to impacts on each logit. Logits are transformed by subtracting the mean of the logits and passing through a ReLU, allowing the evaluator to focus on where logits are raised. Compute the ratio of the amount it raises the logits of the token being attended to, to that of all tokens in this sample. This value ranges from 0 (only raises other tokens) to 0.5 (only raises the present token), so we scale it into the range of -1 to 1.

- *Prefix matching:* Generate a sequence of 25 random tokens, excluding the most common and the least common tokens. Repeat this sequence 4 times and prepend a "start of sequence" token.[31] Compute the attention pattern. The prefix matching score is the average of all attention pattern entries attending from a given token back to the tokens that preceded the same token in earlier repeats.

- *Previous token:* Draw an example from the training distribution. Compute the attention pattern. The previous token score is the average of the entries of all the attention pattern entries attending from token *i* to token *i-1*.

In the Appendix we validate the connection between these empirical measurements and the mathematical properties of attention head weight matrices explored in Transformer Circuits, and show a joint distribution of scores for copying and prefix matching, for heads in the small and large models.

**(4) Trace of QK eigenvalues (for previous-token QK-circuit term)**

In addition to the heuristic measurements described above, we construct another measure of **prefix matching** that draws on concepts from A Mathematical Framework for Transformer Circuits. In short, we measure how overall large and positive the eigenvalues are of the $W$ matrix found in the term of the QK circuit that corresponds specifically to "pure K-composition" with heads that attend to the previous token. Positive eigenvalues of $W$ indicate "same-matching": preferring that this component of the query and key vectors be the same. A "same-matching" $W$ matrix associated with a "previous token" term in the QK circuit is a simple way to implement basic prefix matching behavior. We unpack here more about what this measure is and how it is computed.

The "QK circuit" is a term we coined in our previous work to refer specifically to the entire transformation $C_{QK}^h$ which is applied to the one-hot token vectors $t$ to compute the attention pattern $A^h$ for a specific head: $A^h$ = $\text{softmax}^* \left( t^T \cdot C_{QK}^h t \right)$ . The full computation of $C_{QK}^h$ for a given head can be decomposed into many terms, each corresponding to one of the many paths through the model that needs to be computed in order to map tokens to their attention scores for that head. The section Path Expansion of Attention Scores QK Circuit in our earlier work shows all the terms of a full expansion for a two-layer model; models with more layers have QK circuits with even more terms.

For this measure, we narrow the scope of our calculation to the simplest possible term of the QK circuit that could contribute to the "prefix-matching" behavior of an induction head: namely, only pure K-composition terms (in which the query-side term is just the direct path from the embeddings), and furthermore only the subset of those that correspond specifically to K-composing with an earlier head that consistently attends to the immediately previous token.

In order to compute this measure for a given head, we do the following:

1. For all attention heads in earlier layers, measure how "previous-token-like" the head is, as the average of the attention pattern along the previous-token off-diagonal, over an example text.

2. Compute the $W$ matrix found in each individual $Id \otimes A^{h_k} \otimes W_E^T W_{QK}^h W_{OV}^{h_k} W_E$ term, for each head in earlier layers.

3. Create a combined $W$ matrix by weighting the matrices in step 2 by the measurement in step 1 of how "previous-token-like" each previous head is.

4. Measure the trace (i.e. the sum of the eigenvalues) of this combined weighted $W$ matrix.

Because this measure relies on K-composition with heads in previous layers, it is only defined for models more than one layer deep, and only for attention heads in the second layer and beyond.

We show in the Appendix that this measure correlates well with our heuristic prefix matching evaluator on our small models.

## Analyses

**Per-token losses**

The second through fourth rows of the Model Analysis Table show results derived from the per-token losses from the small models and full-scale models. We show the following "views" on the per-token loss data:

- **Loss curve**: The random-token-per-example loss data — a matrix of size (N model snapshots x 10,000 examples) for N = 200 snapshots per small model and N = 15 snapshots per full-scale model — is averaged over the example dimension, giving an average loss per model snapshot.

- **2D in-context learning plot**: The context-index-average loss data — a matrix of size (N model snapshots x 512 token index positions) — is visualized directly as a 2D plot. This shows model performance on tokens earlier versus later in the context, as well as across training time. We borrow this technique for highlighting "in-context learning" from Kaplan et al. [13] . A horizontal slice of this plot represents a loss curve as a function of training step, for one particular token index rather than averaged over all tokens; a vertical slice represents a particular snapshot's prediction ability as a function of token index.

- **2D loss derivative plot:** We also plot the partial derivative of the context-index-average loss data with respect to log(context index). This captures the amount that token loss is reduced by seeing ϵ% more context.

- **In-context learning score**: We compute what we call the "in-context learning score" as the average loss of the 500th token minus the average loss of the 50th token (for a length-512 context)± over training. This functions as a summary of the 2D in-context learning plot, capturing in a single summary statistic the trajectory over training time of the model's ability to make better predictions later in the context than earlier.

Additionally in a separate row at the top, we show PCA results from the per-token results. These are displayed separately because, unlike all the other graphs, the x-axis is not elapsed training tokens.

- **PCA:** A plot of each model's trajectory in function space over training time, using PCA. The random-token-per-example loss data for all snapshots and all models of each type is concatenated into a large matrix, of shape (N*M model snapshots x 10,000 examples), where N = 200 snapshots per model and M = 12 models for small models, and N = 15 snapshots per model and M = 6 models for full-scale models. This combined matrix is projected down to the first two principal components. Each model is then plotted separately, tracing out a trajectory in function space over the course of its training

**Attention head measurements**

The next two rows of Figure 1 show results derived from the [attention head measurements](#): one row shows the score of the "prefix matching" head activation evaluator over training time, and the following row shows the prefix-matching QK-circuit eigenvalue trace over training time.

For the small models, all attention heads are shown. For the full-scale models, a subset of 100 heads is shown; see the [Appendix](#) for an explanation of how the 100 heads were chosen.

Note that in the QK-circuit plot, the change in weight decay at approximately 5 billion tokens (described in [Model Details](#)) can be seen particularly prominently as a bend or hiccup in the plotted trajectories, but occurs after the phase change and is an unrelated phenomenon.

**Ablation attribution to phase change**

For the small models only, the final two rows of Figure 1 show results derived from the per-token loss data from the attention head ablation experiments. We show only the [pattern-preserving ablation](#) results here; *full* ablation results can be found in the appendix. Note also that ablation experiments were not conducted for the full-scale models.

The first "view" on the ablation data is the attribution to the *"before-and-after" vector*, which relates the result of each head ablation to the change in model behavior over the course of the phase change, and which we explain here. First: For a given model, a length-10,000 vector is computed by taking the delta of the random-token-per-example loss data for the "phase change end" snapshot, minus the "phase change start" snapshot. This represents how the model's behavior changed over the course of the phase change. This vector is normalized to a unit vector; we call this the *before-and-after vector*. It can be thought of as the "direction in token-loss space" that is affected by the phase change. Second: Note that ablating a given head also induces a change in the behavior of a model snapshot, and thus a corresponding length-10,000 vector change to the random-token-per-example loss data (calculated by subtracting the un-ablated snapshot's per-token loss vector from the ablated snapshot's per-token loss vector). We are interested in the similarity between these two changes, which we compute as the dot product between the model's before-and-after vector and the ablation change vector for this head and this snapshot. For induction heads, this dot product is negative, indicating that removing induction heads is like *un*doing the changes to model behavior that happened during the phase change.

The second "view" on the ablation data is the attribution to the *in-context learning score*. We operationalize the model's in-context learning abilities as how much better the model is at predicting the 500th token than the 50th token: this difference measured as Loss($n$train, $i$context= 500) - Loss($n$train, $i$context= 50) is the *in-context learning score*.